\documentclass{article}
\pdfoutput=1 
\usepackage{arxiv}
\usepackage[utf8]{inputenc} 
\usepackage[T1]{fontenc}    
\usepackage{url}            
\usepackage{booktabs}       
\usepackage{amsfonts}       
\usepackage{nicefrac}       
\usepackage{microtype}      
\usepackage{lipsum}		
\usepackage{natbib}

\usepackage{amsmath}
\usepackage{multirow}
\usepackage{amsopn}
\usepackage{caption}
\usepackage{pgfplots}
\usepackage[mathcal]{euscript}
\usepackage{standalone}
\usepackage{multirow}
\usepackage{booktabs}
\usepackage{subfig}
\usepackage{amsfonts}
\usepackage{amssymb}
\usepackage{mathrsfs}

\DeclareMathOperator{\R}{\mathbb{R}}
\pgfmathdeclarefunction{gauss}{2}{%
  \pgfmathparse{1/(#2*sqrt(2*pi))*exp(-((x-#1)^2)/(2*#2^2))}%
}

\DeclareMathOperator{\tscore}{\mathsf{t-score}}

\DeclareMathOperator{\pmi}{\mathsf{PMI}}
\DeclareMathOperator{\npmi}{\mathsf{NPMI}}
\DeclareMathOperator{\pat}{{\it p}\text{@}}
\DeclareMathOperator{\llr}{\mathsf{G^2}}
\DeclareMathOperator{\chis}{\chi^2}
\DeclareMathOperator{\dice}{\mathsf{Dice}}

\DeclareMathOperator{\mOne}{\mathsf{SDMA_m}}
\DeclareMathOperator{\mTwo}{\mathsf{SDMA_h}}
\DeclareMathOperator{\mThree}{\mathsf{SDMA_c}}

\newcommand{\ams}{{AMs}}

\DeclareMathOperator{\compErrorInt}{\mathsf{INTRCT}}
\DeclareMathOperator{\additive}{\mathsf{ADT}}
\DeclareMathOperator{\noncompmult}{\mathsf{MLT}}
\DeclareMathOperator{\ncscore}{\mathsf{MMN}}
\DeclareMathOperator{\ncscoreSMOO}{\mathsf{MMN_s}}

\newcommand{\dsfar}[1]{\textsc{ds\_farahmand}}
\newcommand{\dsred}[1]{\textsc{ds\_reddy}}

\DeclareFontFamily{OT1}{pzc}{}
\DeclareFontShape{OT1}{pzc}{m}{it}{<-> s * [1.10] pzcmi7t}{}
\DeclareMathAlphabet{\mathpzc}{OT1}{pzc}{m}{it}

\title{A Multivariate Model for Representing Semantic Non-compositionality}

\author{
  Meghdad Farahmand\thanks{This research was partly published as part of the PhD dissertation of the author, presented to the Computer Science department of the University of Geneva in March 2017. The author would like to thank Dr. James Henderson for his valuable input throughout this work.}\\
  Department of Computer Science\\
  University of Geneva\\
  \texttt{meghdad.farahmand@gmail.com} \\
}
\begin{document}
\maketitle
\begin{abstract}
  
Semantically non-compositional phrases constitute an intriguing research topic in Natural Language Processing. Semantic non-compositionality --the situation when the meaning of a phrase cannot be derived from the meaning of its components, is the main characteristic of such phrases, however, they bear other characteristics such as high statistical association and non-substitutability. In this work, we present a model for identifying non-compositional phrases that takes into account all of these characteristics. We show that the presented model remarkably outperforms the existing models of identifying non-compositional phrases that mostly focus only on one of these characteristics.

\end{abstract}

\section{Introduction}

Non-compositional phrases are those phrases the meaning of which cannot be directly derived from the meaning of their components, as in {\it soap opera}, {\it kangaroo court}, and {\it ret tape}. Non-compositional phrases are considered to be one of the most important sub-categories of Multiword Expressions (MWEs) and efficient identification of these phrases can have a major impact on semantic applications such as Natural Language Understanding, Sentiment Analysis, Natural Language Generation and Opinion Mining \citep{berend2011opinion}. 

{\em Semantic non-compositionality} (also referred to as {\em semantic idiosyncrasy}) is the main characteristic of non-compositional phrases, however, they have other properties such as {\em high statistical association} and {\em non-substitutability}. In most studies of non-compositional phrases, however, only semantic non-compositionality is considered while statistical association and non-substitutability are widely neglected. Most previous work first specify a distributional representation of words and phrases (e.g. traditional distributional representations or word embeddings); further, they study different functions, in order to compose the representations of words into phrases. Phrases for which the distributional representation of the phrase is considerably different from the representation generated from the representation of their components by the composition function are then regarded as non-compositional \citep{mitchell2008vector,reddy2011empirical,salehi2015embed}. A small fraction of previous work such as  \citet{kiela2013detecting} and \citet{Lin:1999}, on the other hand, try to identify non-compositional phrases through non-substitutability and statistical association which are general properties of all types of MWEs and are not specific to non-compositional phrases. 

The models that attempt to identify non-compositional phrases {\em only} based on their non-compositionality generally have a high precision and a low recall, while those that attempt to identify non-compositional phrases through general properties of MWEs such as statistical association and non-substitutability, have a low precision and a high recall. In this work, we present a multivariate model to identify non-compositional phrases based on their specific and general characteristics which leads to a high precision and recall. To the best of our knowledge, no other work in the literature takes into account both specific and general properties of non-compositional MWEs for their identification and hence unlike the presented model, most available models suffer from either a low precision or a low recall. 

\section{Related Work}
\label{sec:related:work}

Some of the earliest works on non-compositional MWEs include \citet{Tapanainen:1998} who propose a method to identify non-compositional verb-object collocations\footnote{The term collocation refers to the statistically idiosyncratic MWEs in recent work \citep{baldwin2010multiword}. In early work, however, it referred to all types of MWEs.} based on the semantic asymmetry of verb-object relation and \citet{Lin:1999} which was discussed earlier. \citet{baldwin2003empirical} present a method that decides about the non-compositionality of English noun compounds and verb-particle constructions by comparing the vectors of their components against the vector of the phrase. They create the word vectors by means of Latent Semantic Analysis (LSA). \citet{McCarthy03detectinga} devise a number of measures for non-compositionality based on the comparison of the neighbors of phrasal verbs and their corresponding simplex verbs. \citet{venkatapathy:2005:measuring:VN} present a supervised model that ranks the MWE candidates based on their non-compositionality. \citet{Katz:2006} test whether the local context of an MWE can distinguish its idiomatic use from its literal use. \citet{reddy2011empirical} employ the additive and multiplicative composition functions of \citet{mitchell2008vector} and several similarity-based models to measure the compositionality of noun compounds. \citet{hermann2012unsupervised} present a model that compares the distributional vectors of a compound and its components and decides about the semantic contributions of different components and subsequently the lexicality of the compound. \citet{im2013exploring} employ various word vector models to decide about the non-compositionality of German noun compounds. They show that window-based models of distributional semantics outperform the syntax-based models in identifying non-compositionality. \citet{kiela2013detecting} present a model of detecting non-compositionality based on the hypothesis that the average distance between a phrase vector and its alternative phrase (created by substituting the components of the original phrase with their similar words) vectors is related to its compositionality. Their models show a small improvement ($+0.014$ and $+0.007$) over their baselines. The models discussed so far are based on traditional vector representations and predefined composition functions. More recent work on non-compositionality, however, rely on word embeddings \citep{salehi2015embed}, and more complex composition functions \citep{yazdaniFarahmand15}. An evaluation of a variety of models that are based on distributional semantics, and the effect of their hyper-parameters on predicting the compositionality of noun compounds in French and English is carried out by \citet{P16-1187}. 

\section{Method}
 
In Sec. \ref{sec:data} and \ref{sec:eval:method}, we present the datasets and the evaluation measures that were used in this research. In Sec. \ref{sec:identify:different:components}, we discuss general and specific characteristics of non-compositional phrases i.e. statistical association, non-substitutability and non-compositionality and present different models to independently identify these characteristics. Finally, in Sec. \ref{sec:mult:baseline} and \ref{dist:of:noncomp} we present a multiplicative baseline and a multivariate-distribution-based model that considers all of these characteristics in order to identify non-compositional phrases. 

\subsection{Data}
\label{sec:data}
We focus on noun-noun compounds due to their high frequency and availability of the respective data sets. 
We use the datasets of  \citet{farahmand2015data} and \citet{reddy2011empirical} (\dsfar{}  and \dsred{} hereafter). 

\dsfar{} contains $1042$ English compounds judged independently by four experts for their statistical idiosyncrasy (their components have a high statistical association) and non-compositionality. \dsred{} contains $90$ compounds judged for their compositionality by crowdsourcing through Amazon Mechanical Turk. See Sec. \ref{multivar:mul:model:eval} for a detailed analysis of these datasets.

While in this work we focus mainly on noun-noun compounds, the presented models can be applied to other syntactic categories of MWEs, such as verb-object combinations and higher order noun compounds with some adjustment.

\subsection{Evaluation Measures}
\label{sec:eval:method}

To evaluate the presented models, we use precision at $k$ ($\pat k$) and Spearman's $\rho$ correlation. $\pat k$ is equivalent to the precision graphs of {\it n-best lists} method of \citet{evert2005statistics} that are commonly used to evaluate the quality of MWE extraction models (for further discussion on the advantages of evaluating the identification of MWEs in this fashion, see \citet{evert2005statistics-patk}). 

\subsection{General and Specific Characteristics of Non-compositional phrases}
\label{sec:identify:different:components}

In Sec. \ref{sec:identify:non-comp}, we present models for identifying semantic non-compositionality (the specific characteristic of non-compositional phrases). To identify the statistical association and non-substitutability (general characteristics of non-compositional phrases), we present several Association Measures (\ams{}) in Sec. \ref{sec:identify:association}, and in Sec \ref{sec:nonsub} we develop a ratio for measuring non-substitutability. 

\subsubsection{Identifying Non-compositionality}
\label{sec:identify:non-comp}

In order to measure semantic non-compositionality (the specific characteristic of non-compositional phrases), we choose the additive model of \citet{reddy2011empirical} adapted to word embeddings \citep{salehi2015embed} ($\additive$ hereafter); and the polynomial regression based model of \citet{yazdaniFarahmand15} with interactive terms ($\compErrorInt$ hereafter), both were shown to outperform the baselines in the related articles. \footnote{\citet{yazdaniFarahmand15} improve the performance of $\compErrorInt$ through lasso regularization, auto-reconstruction and latent annotations. However, since the exact parameter setting for these techniques were not available, we only consider the base form of this model.}

\subsubsection{Identifying the Statistical Association}
\label{sec:identify:association}

As discussed earlier, high statistical association is a general property of non-compositional phrases and it has been commonly recognized through \ams{}. Here, we choose the following six \ams{} that are commonly used to identity the phrases whose components have a high statistical association:

{\bf 1.} $\pmi$ \citep{church1990word} {\bf 2.} normalized $\pmi$ ($\npmi$) \citep{bouma2009normalized} {\bf 3.} $\tscore$ {\bf 4.} Chi-squared ($\chis$) {\bf 5.} Log-likelihood ratio ($\llr$) \citep{dunning1993accurate} {\bf 6.} Dice coefficient ($\dice$) (first applied to MWEs by \citet{smadja1996translating}). 

The above measures were shown to outperform other \ams{} in identifying statistically idiosyncratic phrases \citep{acosta2011identification,bouma2009normalized,evert2005statistics,schone2001knowledge}. Some of these measures, e.g. $\npmi$ (that is an attempt to make the interpretation of $\pmi$ more meaningful by ranging between $0$ to $1$, and less sensitive to low frequency data), $\chis$ and $\llr$ achieve state-of-the-art performance. For a comprehensive study of \ams{} see \citet{evert2005statistics} and \citet{pecina}. 

We evaluate the above \ams{} on \dsfar{} that comes with annotations for statistically idiosyncratic phrases. First, we create a vote-based score from the annotations of this dataset. We give a compound that is annotated as {\em statistically} or {\em semantically} idiosyncratic by one judge, score 1, a compound that is annotated as such by two judges score 2 and so on. 

\begin{figure}[ht!]
\centering
\begin{tikzpicture}
\pgfplotsset{grid style={help lines}}
\begin{axis}[
xmax=400,xmin=0,
ymax=0.95,ymin=0.6,
width=10cm,
height=6.5cm,
legend style={at={(1,1)},anchor=north east},
   xlabel={$k$},
   ylabel={$\pat k$},
   ylabel style = {font=\small,yshift=-2ex},
   yticklabel style = {font=\small,xshift=0.5ex},
   xticklabel style = {font=\small,yshift=0.5ex},
grid=both,
grid style={line width=.1pt, draw=gray!10},
major grid style={line width=.2pt,draw=gray!50},
minor tick num=1,
legend entries={$\pmi$,$\npmi$,$\tscore$,$\chis$,$\llr$,$\dice$},
]
\addplot[color=red] table {data/ch_statistical_idiosyn/FAR_J2/baselines/testIndex/k_pies_pmi.csv};
\addplot[color=blue,style=dashdotted,mark=star] table {data/ch_statistical_idiosyn/FAR_J2/baselines/testIndex/k_pies_npmi.csv};
\addplot[color=green,style=dashdotted,mark=star] table {data/ch_statistical_idiosyn/FAR_J2/baselines/testIndex/k_pies_t.csv};
\addplot[color=magenta,style=dashdotted,mark=star] table {data/ch_statistical_idiosyn/FAR_J2/baselines/testIndex/k_pies_chi.csv};
\addplot[color=orange,mark=triangle] table {data/ch_statistical_idiosyn/FAR_J2/baselines/testIndex/k_pies_llr.csv};
\addplot[color=teal,style=dashdotted,mark=star] table {data/ch_statistical_idiosyn/FAR_J2/baselines/testIndex/k_pies_dice.csv};
\end{axis}
\end{tikzpicture}
\vspace{-1.5mm}
\caption{Performance of \ams{} in terms of $\pat k$ on \dsfar{}.}
\label{fig:all:ams:pat:dsfar}
\end{figure}
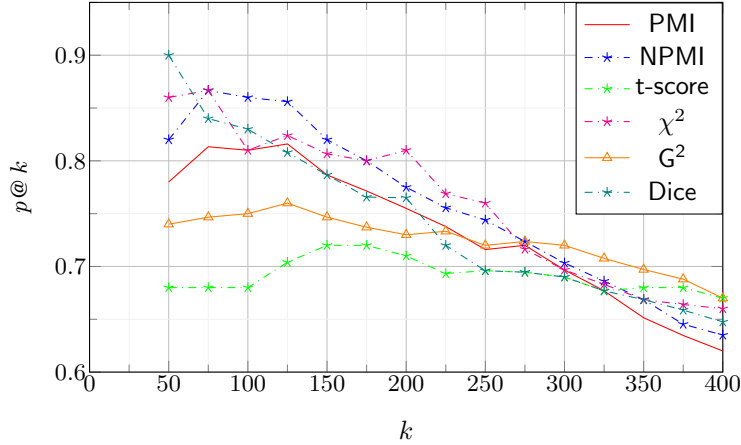

Since the majority of the compounds of this dataset are not idiosyncratic at any level, Spearman's $\rho$ correlation between the models' scores and human judgments is not applicable. Hence, we use only $\pat k$ for the evaluations on this dataset. To measure $\pat k$, we assume any compound that has a human score of $\geq 2$ (annotated as idiosyncratic by at least two judges) is actually idiosyncratic and regard it as a positive instance in our evaluation. We then rank the compounds using the described \ams{} and measure their $\pat k$ for different values of $k$. The results are shown in Fig. \ref{fig:all:ams:pat:dsfar}. As seen, $\chis$ and $\npmi$ generally perform better than other measures, hence, we keep them as baseline for identifying the statistical association and later integrate them in the downstream multiplicative and multivariate models of identifying non-compositional phrases.

\subsubsection{Identifying Non-substitutability}
\label{sec:nonsub}

Similar to the statistical association, non-substitutability is another general property of non-compositional phrases. Non-substitutability means that the components of an idiosyncratic phrase cannot be replaced with their synonyms \citep{manning1999foundations}. Non-substitutability has been discussed in various works as a salient characteristic of MWEs \citep{manning1999foundations,schone2001knowledge,pearce2001synonymy,baldwin2003empirical,ramisch2012generic}, however, unlike statistical association, it is not well-studied especially from a computational perspective except in a few works that propose models that are limited and labor-intensive \citep{pearce2001synonymy} or computationally expensive \citep{frahmand2016loglinear}. In the following, we propose a ratio for measuring non-substitutability that we refer to as Substitution-driven Measure of Association (SDMA). It can be thought of as a measure of statistical association that takes into account the degree of semantic non-substitutability of the phrase to which it is applied. We develop three variations of SDMA and evaluate them in comparison with the best \ams{} from Sec. \ref{sec:identify:association} as well as non-substitutability models of \citet{frahmand2016loglinear}. It turns out that SDMAs achieve a considerably higher performance than previous work and \ams{} in identifying general idiosyncrasy. 

Let us first define the {\em probability of alternatives} as the probability of alternative compounds for the compound $w_1w_2$. We define this probability in three different ways which we refer to as $p_m$ (alternatives generated by substituting the modifier\footnote{Since our focus is on English noun compounds and the majority of these compounds are right-headed, we refer to the left (first) word as the head and the right (second) word as the modifier.}), $p_h$ (alternatives generated by substituting the head) and $p_c$ (alternatives generated substituting the head and the modifier):
\begin{equation}
p_m(w_1,w_2) =\sum\limits_{w'_1 \in \mathpzc{S}_{w_1}} \frac{C(w'_1,w_2)+1 }{N+\mathcal{L}}
\end{equation}

where $\mathpzc{S}_{w_1}$ is the set of $k$ nearest neighbors to $w_1$ in the word vector space. We use  {\it fastText} $300d$ Wikipedia vectors. We set $k$ equal to $5$ after experiencing with different values. We observed that  $k=3$ leads to missing some plausible semantically related words and $k=7$ leads to the inclusion of irrelevant words. 
$N$ is the number of all word pairs in the corpus and $C(w_1, w_2)$ is the number of times that compound $w_1w_2$ appeared in the corpus. $\mathcal{L}$ in the denominator and addition by one in the numerator represent a Laplace smoothing with parameter $1$. Note that in order to ensure a well-defined distribution where $\sum_{w_1,w_2 \in T} p_m(w_1,w_2) = 1$, a hard clustering must be applied to ensure that words do not appear in more than one semantic cluster. Analogously, we can define $p_h$ and $p_c$ as follows:

\begin{equation}
p_h(w_1,w_2) =\sum\limits_{w'_2 \in \mathpzc{S}_{w_2}} \frac{C(w_1,w'_2)+1 }{N+\mathcal{L}}
\end{equation}
 
\begin{equation}
p_c(w_1,w_2) =\sum\limits_{w'_1 \in \mathpzc{S}_{w_1}} \sum\limits_{w'_2 \in \mathpzc{S}_{w_2}} \frac{C(w_1',w'_2)+1 }{N+\mathcal{L}}
\end{equation}

In summary, $p_m$, $p_h$, and $p_c$ are defined as the sum of the probabilities of those bigrams that can semantically substitute $w_1w_2$. For example, for the compound {\it weather forecast}, substitution is defined by $p_m$ as substitution of the modifier (e.g. {\it climate forecast}), substitution of the head by $p_h$ (e.g. {\it weather prediction}), and substitution of both constituents by $p_c$ (e.g. {\it climate prediction}). Subsequently, we define SDMAs as follows:

\begin{equation}
\mOne(w_1,w_2) = \log \frac{p(w_1,w_2)}{p_m(w_1,w_2)}
\end{equation}

SDMAs, in simple terms, are equal to the log of the joint probability of a word pair reduced by a factor of the probability of alternative pairs. $\mOne$ assumes that the probability of alternatives is defined by $p_m$. The next two variations, i.e. $\mTwo$ and $\mThree$ are defined in the same way but they assume that the probability of alternatives is defined by $p_h$ and $p_c$, respectively. 

\begin{equation}
\mTwo(w_1,w_2) = \log \frac{p(w_1,w_2)}{p_h(w_1,w_2)}
\end{equation}

\begin{equation}
\mThree(w_1,w_2) = \log \frac{p(w_1,w_2)}{p_c(w_1,w_2)}
\end{equation}

Advantages of measuring non-substitutability in this fashion are {\em low computational costs}, {\em wide coverage}, and {\em no need for human involvement} that can be costly and slow. Previous work on modeling non-substitutability lack one or more of the above.\\

\subsubsection*{Evaluation of SDMAs}
To evaluate SDMAs, we follow the same method described for the evaluation of \ams{} in Sec. \ref{sec:identify:association}. We use the discussed \ams{} and log-linear based non-substitutability models of \citet{frahmand2016loglinear} ($H_1$ and $H_2$). 

\begin{figure}[ht!]
\centering
\begin{tikzpicture}

\pgfplotsset{grid style={help lines}}
\begin{axis}[
xmax=400,xmin=0,
ymax=1,ymin=0.65,
width=10cm,
height=6.5cm,
legend style={at={(0.75,0.82)},anchor=west},
   xlabel={$k$},
   ylabel={$\pat k$},
   ylabel style = {font=\small,yshift=-2ex},
   yticklabel style = {font=\small,xshift=0.5ex},
   xticklabel style = {font=\small,yshift=0.5ex},
grid=both,
grid style={line width=.1pt, draw=gray!10},
major grid style={line width=.2pt,draw=gray!50},
minor tick num=1,
legend entries={$\mOne$,$\mTwo$,$\mThree$,$H_1$,$H_2$,$\npmi$,$\chis$},
]

\addplot[color=green!60!black,mark=10-pointed star] table {data/ch_statistical_idiosyn/FAR_J2/SDMA/test/D200/k_pies_m1.csv};
\addplot[color=cyan!50!black,mark=asterisk] table {data/ch_statistical_idiosyn/FAR_J2/SDMA/test/D200/k_pies_m2.csv};
\addplot[color=red,mark=x] table {data/ch_statistical_idiosyn/FAR_J2/SDMA/test/D200/k_pies_m3.csv};

\addplot[color=orange,mark=triangle] table {data/ch_statistical_idiosyn/FAR_J2/Hybrid/k_pies_H1.csv};
\addplot[color=blue,mark=triangle] table {data/ch_statistical_idiosyn/FAR_J2/Hybrid/k_pies_H3.csv};

\addplot[color=blue,style=dashdotted,mark=star] table {data/ch_statistical_idiosyn/FAR_J2/baselines/testIndex/k_pies_npmi.csv};
\addplot[color=magenta,style=dashdotted,mark=star] table {data/ch_statistical_idiosyn/FAR_J2/baselines/testIndex/k_pies_chi.csv};
\end{axis}
\end{tikzpicture}
\hspace{10in}\parbox{3in}{\caption{Performance of SDMAs in terms of $\pat k$ in comparison with AMs, $H_1$, and $H_2$ on \dsfar{}.\label{fig:sdmas:pat}}}
\end{figure}
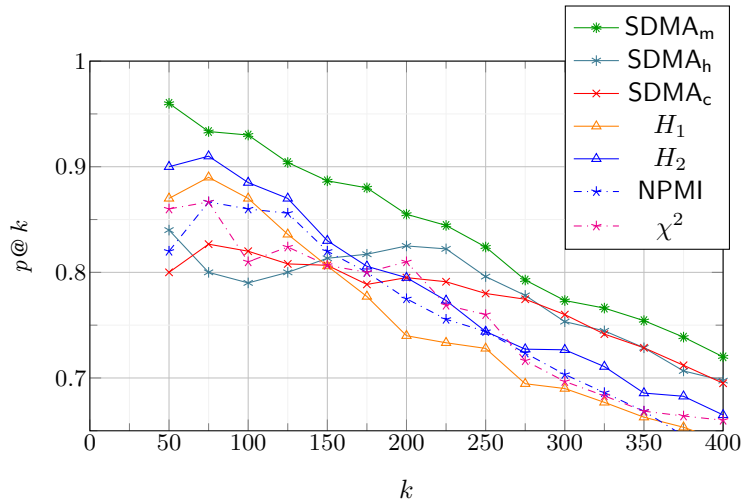

The results are shown in Fig. \ref{fig:sdmas:pat}. As can be seen, $\mOne$ performs considerably better than other models outperforming the best performing \ams{} and best previous non-substitutability measures. $\mTwo$ and $\mThree$ on the other hand, while performing poorer than the baselines for small $k$, at around $k=200$ they gain on the baselines and outperform them thereafter. Differences between performance of SDMAs show the effects of the directionality of noun compounds. Superiority of $\mOne$ shows the important role of the head in shaping the idiosyncrasy and fixedness of a compound. Performing well at higher values of $k$ for $\mTwo$ and $\mThree$ shows the role of the modifier in forming idiosyncratic compounds with less degree of fixedness and consequently a less significant idiosyncrasy. 

\subsection{Hybrid Baseline for Identifying Non-compositionality}
\label{sec:mult:baseline}

In the previous section, we introduced various measures to identify general and specific characteristics of non-compositional phrases. In particular, we presented \ams{} to identify statistical association, SDMAs to identify non-substitutability, and $\additive$ and $\compErrorInt$ to identify non-compositionality. Let us now introduce a simple multiplicative baseline that takes into account all of the above characteristics in order to identify non-compositional phrases. We refer to this baseline as $\noncompmult$. 

To formulate $\noncompmult$, we chose $\compErrorInt$ due to its superior performance from among the non-compositionality models. Analogously, we choose the best performing \ams{} and non-substitutability measures, i.e. $\npmi$ and $\mOne$. We then combine these measures through multiplication:

\begin{equation}
\noncompmult = \compErrorInt \times \mOne \times \npmi
\label{noncom:score:pmult}
\end{equation}

As we will see in the evaluation of $\noncompmult$ in Sec. \ref{multivar:mul:model:eval}, while jointly considering different characteristics of non-compositional phrases through multiplication can improve their identification, this approach has several drawbacks. For instance, a candidate that has a very high $\npmi$ can still get a high $\noncompmult$ score regardless of its non-compositionality score (cf. \citet{kiela2013detecting} and \citet{Lin:1999}). More specifically, compositional MWEs with relatively high values of \ams{} and/or SDMAs and a low degree of non-compositionality can still have a high $\noncompmult$. 

\subsection{Multivariate Distribution-based Model for Identifying Non-compositionality}
\label{dist:of:noncomp}

Non-compositional phrases, in addition to their non-compositionality have the general properties of idiosyncratic phrases. In other words, they are non-compositional, statistically idiosyncratic, and non-substitutable. That means for non-compositional phrases, the corresponding scores of all of these characteristics should have a high value. This implies that non-compositional phrases must appear more densely to the right of the $mean$ of any of the above scores, assuming the distribution of that score is (approximately) normal. 

In addition to the value of different scores, we can take advantage of their probabilities. Meaning that we can estimate the probability of a score and take this probability into account for the identification of a non-compositional phrase. This leads to a remarkable advantage over working with plain values only. 

Let us first look at the distribution of \ams{}, SDMAs, and $\compErrorInt$ on \dsfar{} that can be more representative due its larger size compared to \dsred{} (1042 vs 90). 

\begin{figure*}[ht!]
\centering
    \subfloat[$\npmi$]{\includegraphics[scale=0.23]{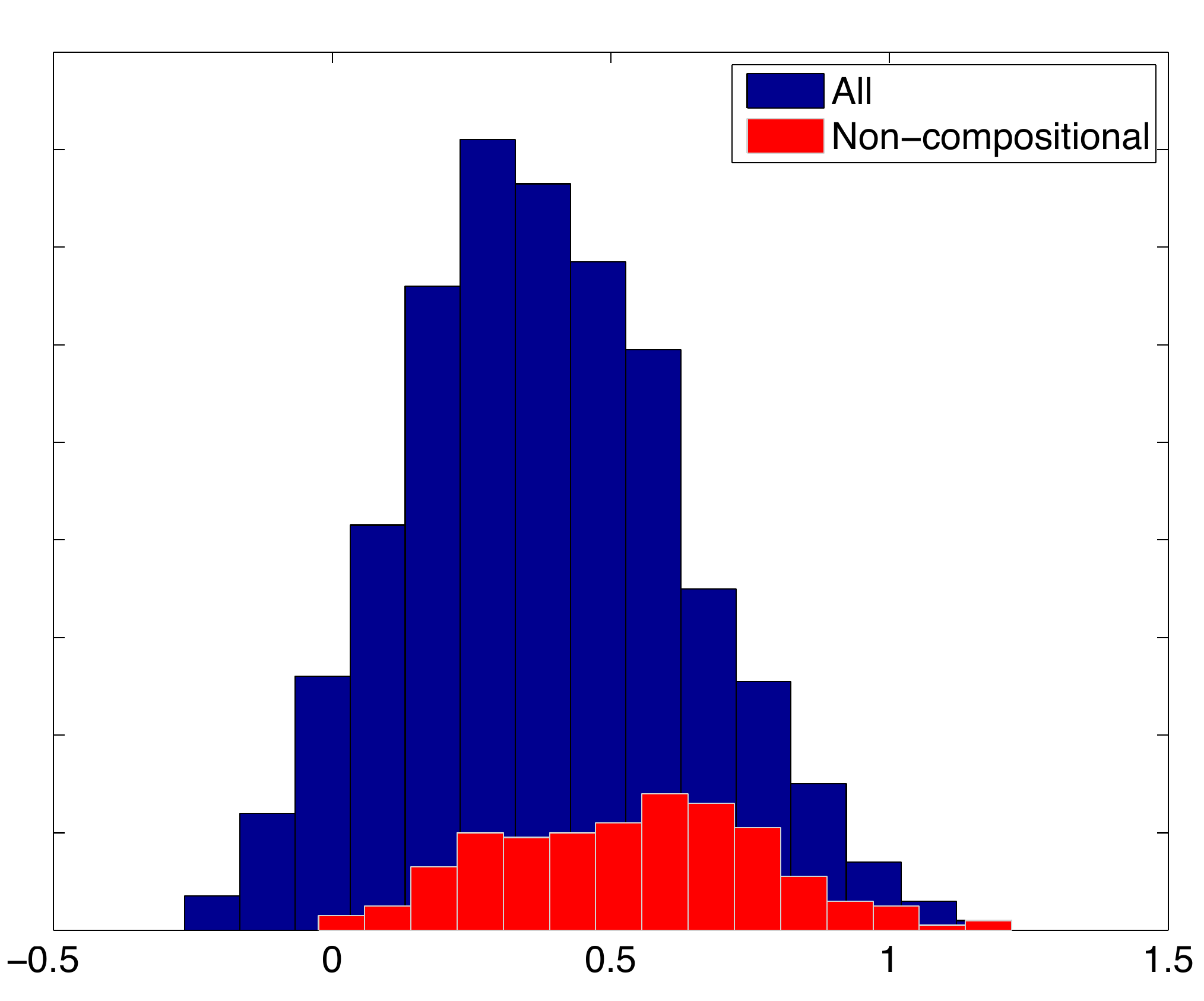}}
    \subfloat[$\chis$]{\includegraphics[scale=0.23]{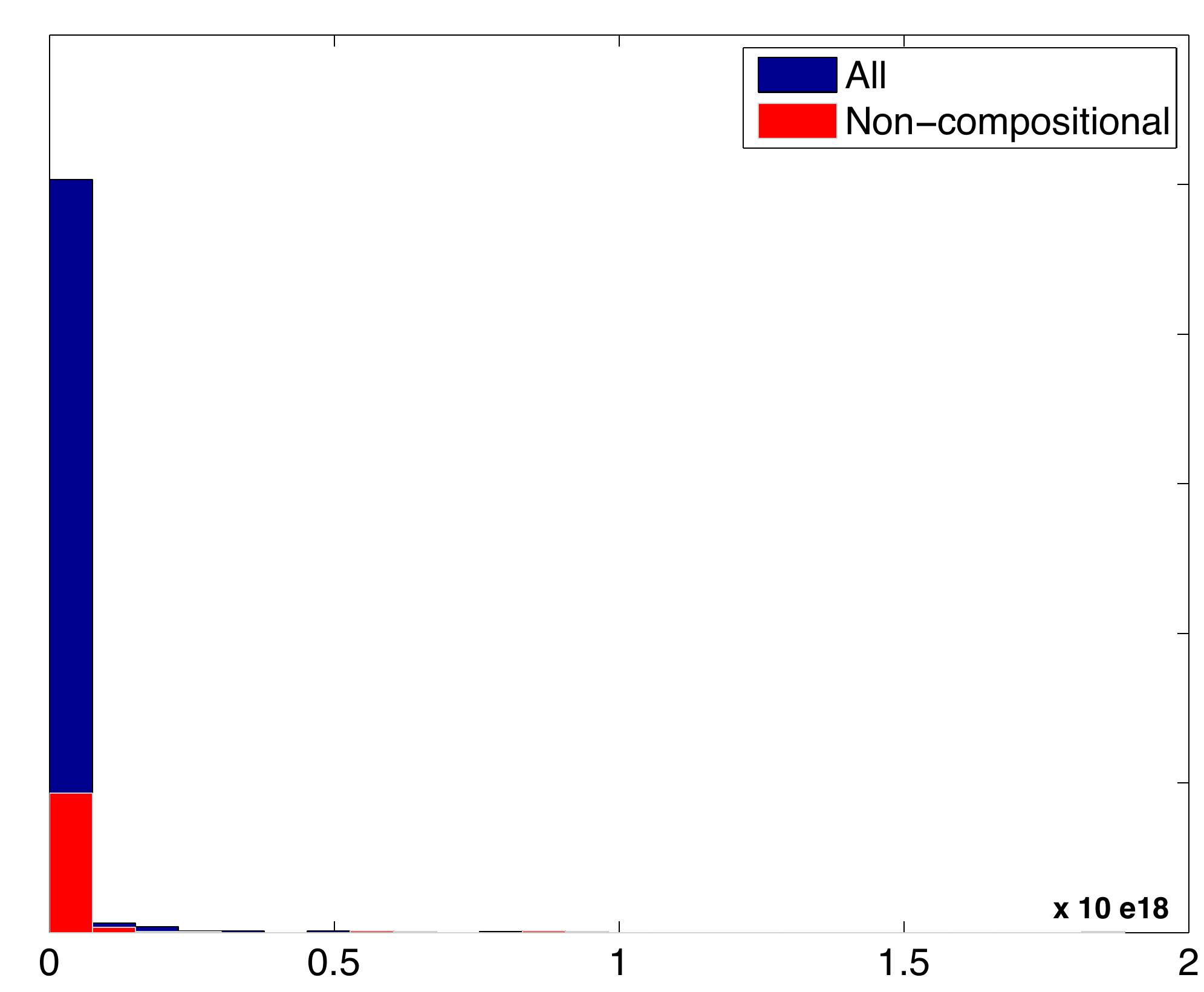}}
\caption{Distributions of $\npmi$ and $\chis$ in \dsfar{}.}
\label{fig:ch:noncomp:distributions:ams}
\end{figure*}

\begin{figure*}[ht!]
\centering
  \subfloat[$\mOne$]{\includegraphics[scale=0.23]{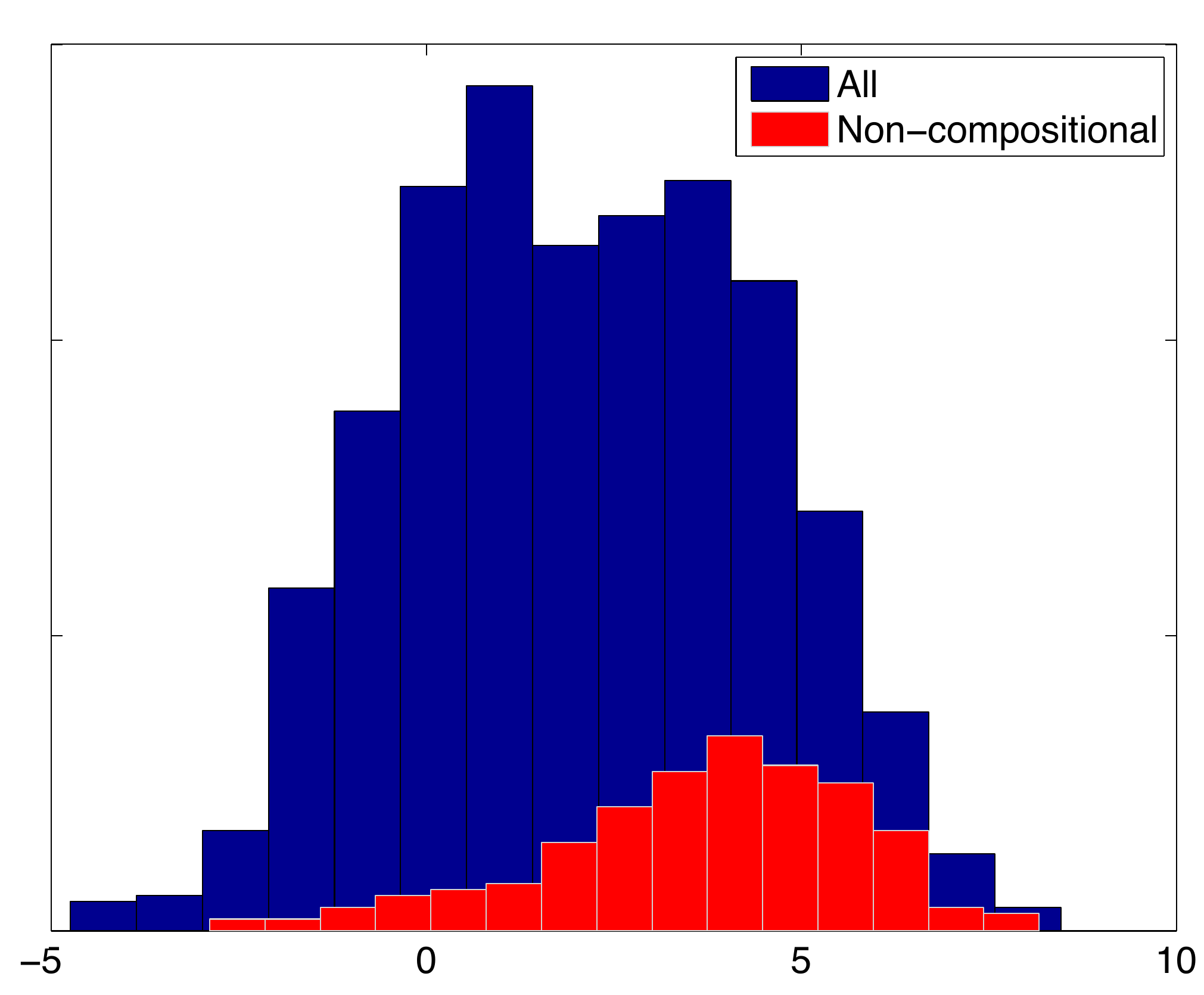}}
    \subfloat[$\mTwo$]{\includegraphics[scale=0.23]{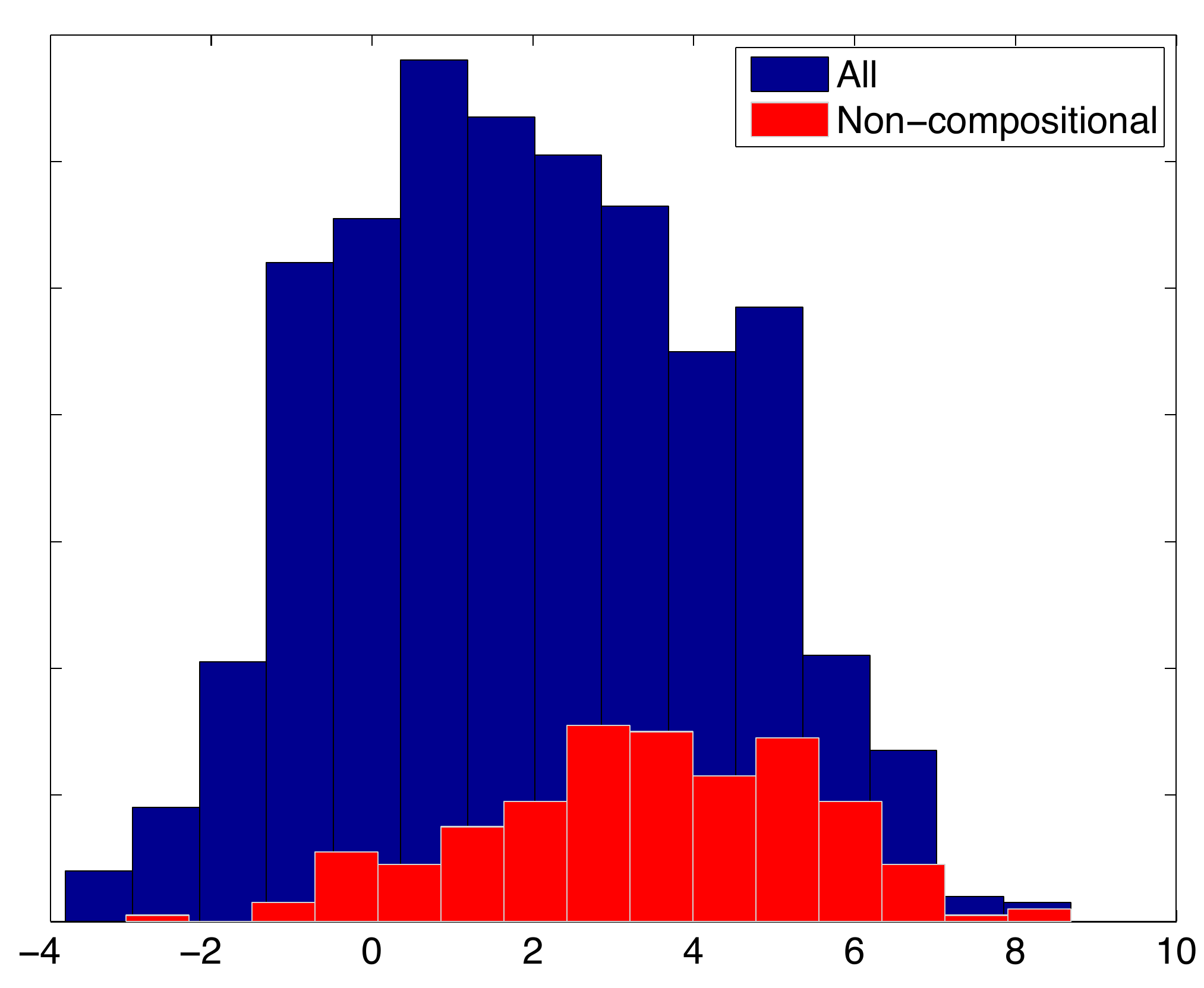}}
    \subfloat[$\mThree$]{\includegraphics[scale=0.23]{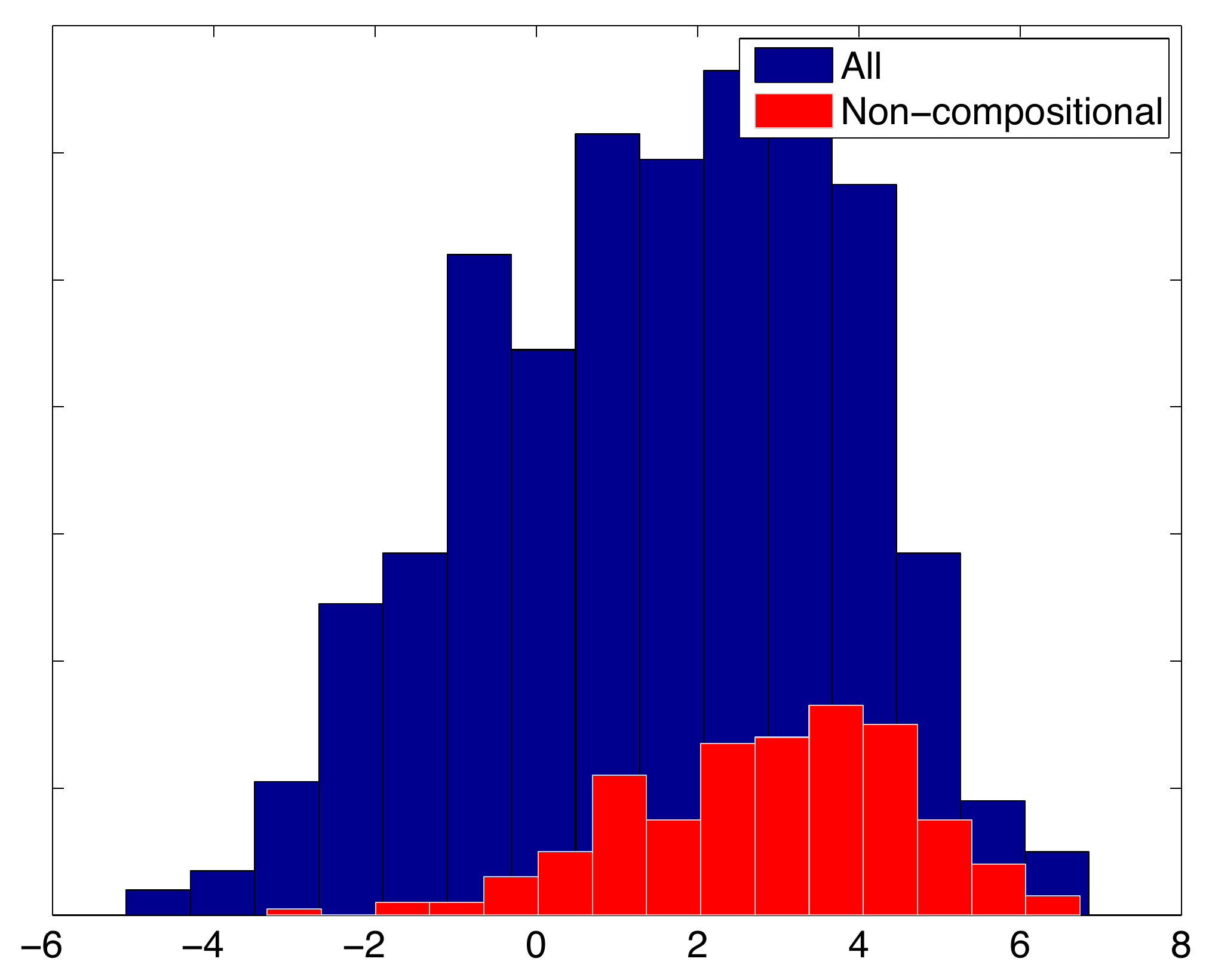}}
    \hfill
\caption{Distributions of SDMAs in \dsfar{}.}
\label{fig:ch:noncomp:distributions:sdmas}
\end{figure*}

The distributions of the best performing \ams{} and SDMAs on \dsfar{} are shown in blue in Fig. \ref{fig:ch:noncomp:distributions:ams} and \ref{fig:ch:noncomp:distributions:sdmas}, respectively. 
While $\npmi$ scores are approximately normally distributed\footnote{Normality of the distributions were tested by Jarque-Bera test. While most of the distributions are not perfectly normal, they are either skewed and hence easily transformable to a normal distribution or near normal and hence can be directly exploited by the proposed model.}, $\chis$, has a spiky distribution that is far form normal. Although it might be possible to transform the distribution of $\chis$ into normal, we do not investigate it any further. $\mOne$, $\mTwo$ and $\mThree$ on the other hand are all near-normally distributed. For every measurement, the fraction of non-compositional instances  (any instance that has a human non-compositionality score of at least 2 out of 4) in each bin is shown in red. As we have already discussed, for almost all scores, most non-compositional instances appear in the right side of their $mean$. Additionally, referring back to the previous discussion about taking the probability of a score into account in addition to its value, we can establish that the lower this probability is, the higher the chance of being non-compositional becomes. In the distribution of a single score, these conditions can be true for many non-idiosyncratic and compositional phrases as well, however if we take into account more scores, each representing a different characteristic of non-compositional phrases, on the resulting multivariate distribution, fewer and fewer compositional phrases satisfy these conditions as we extend this distribution along relevant dimensions. 


\begin{figure*}[ht!]
\centering
  \subfloat[$\compErrorInt$]{\includegraphics[scale=0.23]{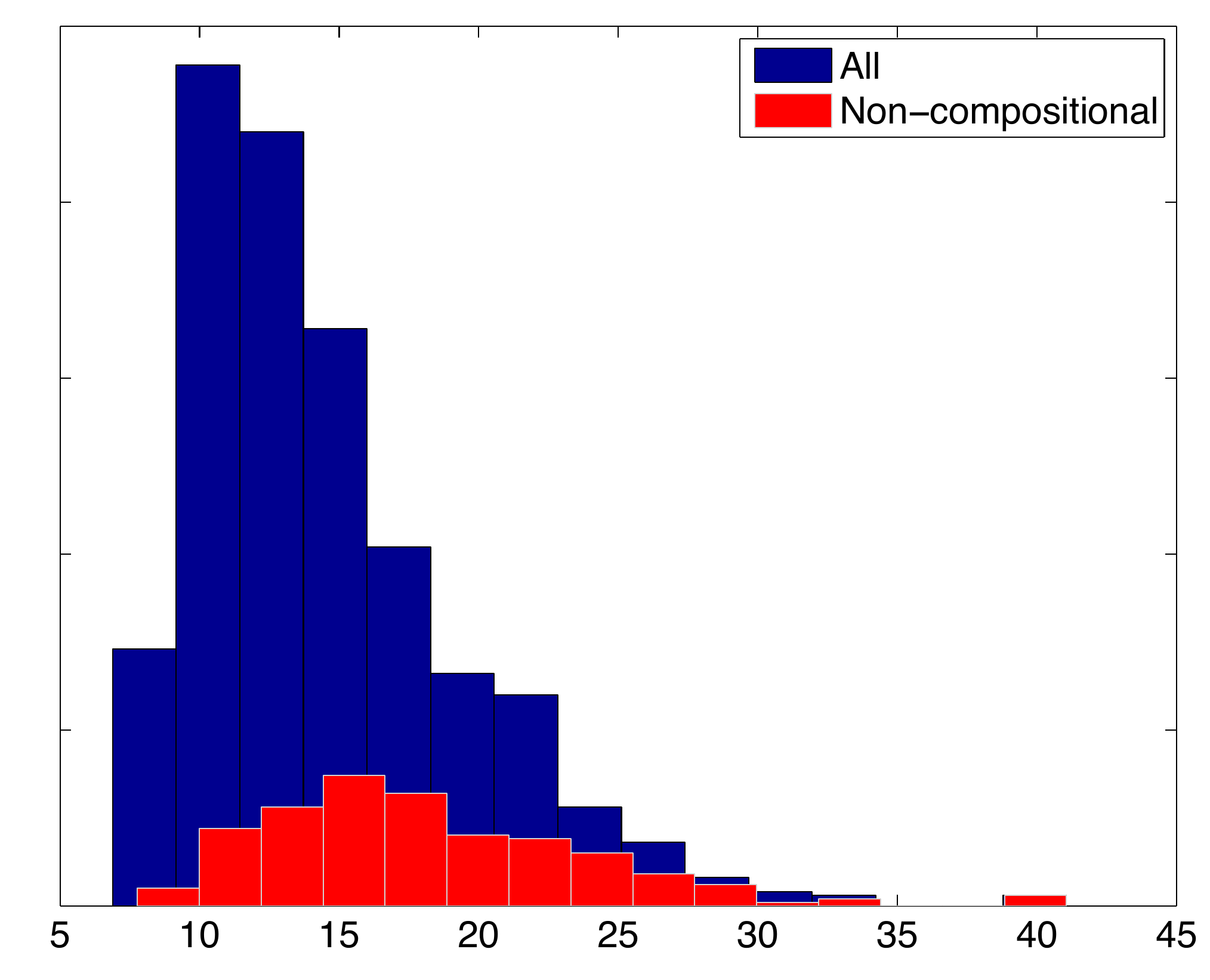}}
  \subfloat[$\log - \compErrorInt$]{\includegraphics[scale=0.23]{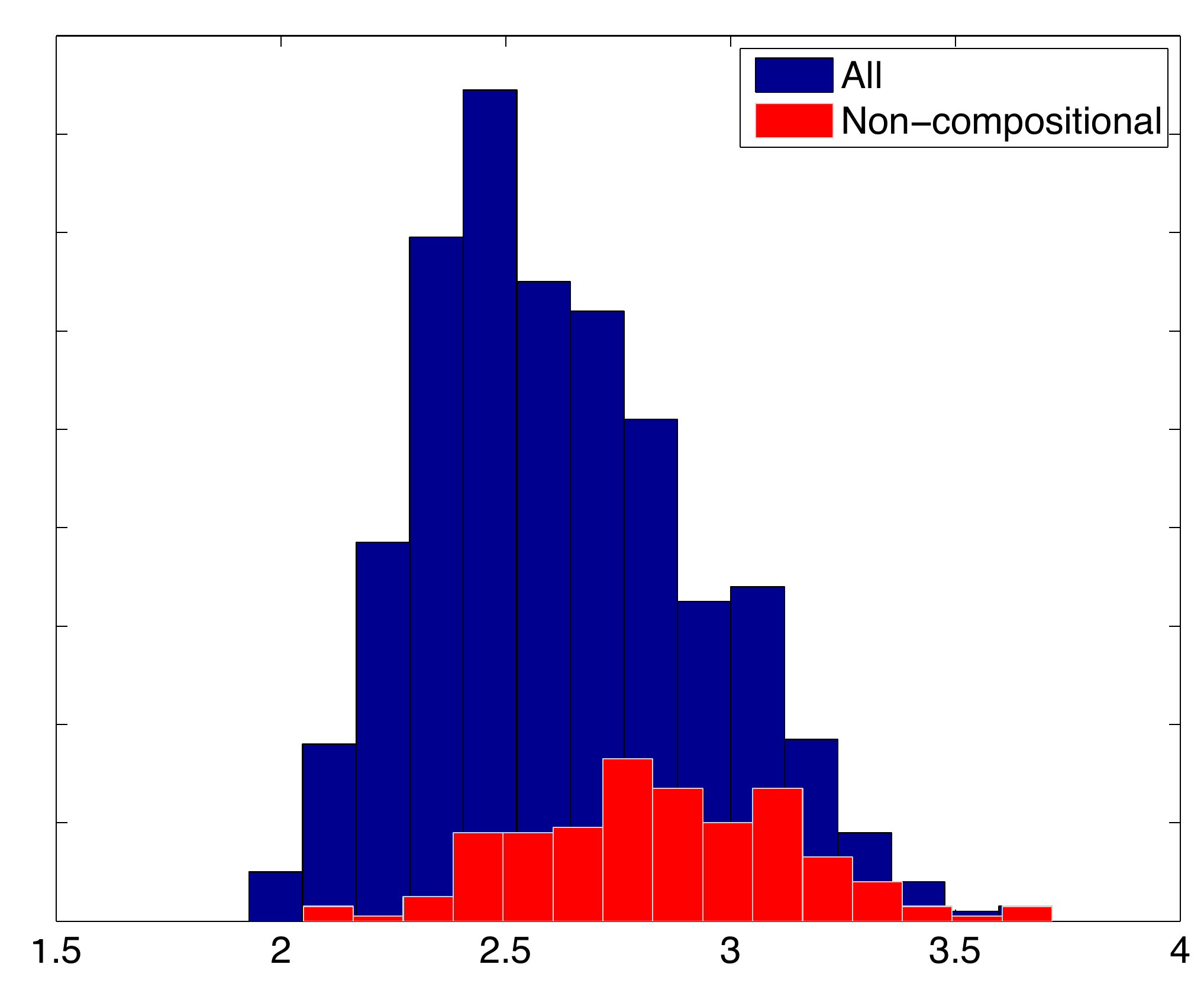}}
\caption{a: Distribution of $\compErrorInt$ in \dsfar{}. b: Distribution of $\log - \compErrorInt$ in \dsfar{}.}
\label{fig:ch:noncomp:distributions:erros}
\end{figure*}

The distribution of $\compErrorInt$ on this dataset can be seen in Fig. \ref{fig:ch:noncomp:distributions:erros}. As seen, this distribution is right skewed, however, it becomes near normal via a log-transformation. In the log-transformed distribution, again non-compositional instances are more densely gathered to the right side of the $mean$. 

\subsubsection{Model Development\protect\footnote{The implementations of the presented models are available at: \protect\url{https://github.com/meghdadFar}}}
\label{multivar:model}

Assume any given phrase $d$ can be represented by $\mathbf{v} \in \R^n$ where each dimension of $\mathbf{v}$ corresponds to one of the discussed scores that represents a general or specific characteristic of non-compositional phrases. If each element of this n-dimensional representation $v_i$ is (near) normally distributed i.e. $v_i \sim \mathcal{N} (\mu_i,\sigma^2_i)$, by making the independence assumption between $v_i$\footnote{The assumption of independence was made because $v_i$ are generated through independent processes.}, a multivariate probability for $d$ can be estimated with respect to the values of its various characteristics as follows:\footnote{If $v_i$ distribution is not perfectly normal for some $i \in n$, the model presented above still works though based on the assumption of normality.}

\begin{equation}
\begin{aligned}
p(\mathbf{v}) & = p(v_1)p(v_2)..p(v_n) \\
& = p(v_1;\mu_1,\sigma_1^2)p(v_2;\mu_2,\sigma_2^2)..p(v_n;\mu_n,\sigma_n^2) \\
& =\prod_{i=1}^{n} p(v_i;\mu_i,\sigma_i^2)
\end{aligned}
\end{equation}
The parameters $\mu_1..\mu_n$ and $\sigma^2_1..\sigma^2_n$ of this multivariate distribution can be estimated by maximum likelihood estimation over $m$ phrases as follows:
\begin{equation}
\mu_i = \frac{1}{m} \sum_{j=1}^m v_i^j
\end{equation}
\begin{equation}
\sigma^2_i = \frac{1}{m} \sum_{j=1}^m (v_i^j-\mu_i)^2
\end{equation}
Having estimated the parameters, the multivariate probability of $d$ represented by $\mathbf{v \in \R^n}$ with respect to $v_i$ can be estimated as follows:

\begin{equation}
\begin{aligned}
p(\mathbf{v}) & = \prod_{i=1}^{n} p(v_i;\mu_i,\sigma_i^2) = \prod_{i=1}^{n} \frac{1}{\sqrt{2\pi\sigma^2_i}} exp\bigg(-\frac{(v_i-\mu_i)^2}{2\sigma^2_i}\bigg)
\end{aligned}
\label{eq:ch:noncomp:pv:expand}
\end{equation}

To identify non-compositional phrases with respect to their characteristics measured by various scores we develop a multivariate score that is in a direct relationship with the value of the characteristic scores -while ensuring that they are greater than $mean$, and in an inverse relationship with the probability of the scores. We refer to this score as Multivariate Measure of Non-compositionality ($\ncscore$) and define it as follows:

\begin{equation}
\ncscore = \prod_{i=1}^n r(v_i-\mu_i)(1-p(\mathbf{v}))
\label{eq:ch:noncomp:nc:score:final}
\end{equation}

$r$ is a rectifier function that maps a negative input to zero while leaving a positive input unchanged. The second term in this equation guarantees an inverse relationship with the probabilities. Rectifier $r$ and smooth rectifier $r_s$ are defined as follows:
\begin{equation}
r(x) = max(0,x)
\label{eq:ch:noncomp:rect}
\end{equation}

\begin{equation}
r_s(x) = \ln (1+e^x)
\label{eq:ch:noncomp:rect:smooth}
\end{equation}

In Eq. \ref{eq:ch:noncomp:nc:score:final}, by replacing $p(\mathbf{v})$ with the right side of Equation \ref{eq:ch:noncomp:pv:expand}, $\ncscore$ can be expanded as follows:

\begin{equation}
\ncscore = \prod_{i=1}^n r(v_i-\mu_i)
\times \left[ 1- \prod_{i=1}^{n} \frac{1}{\sqrt{2\pi\sigma^2_i}} exp\bigg(-\frac{(v_i-\mu_i)^2}{2\sigma^2_i}\bigg) \right]
\label{eq:ch:noncomp:nc:score:expanded}
\end{equation}

The smooth version ($\ncscoreSMOO$) with a smooth rectifier can be defined as follows:
\begin{equation}
\ncscoreSMOO =\prod_{i=1}^n  \ln (1+\exp(v_i-\mu_i))
\times \left[ 1- \prod_{i=1}^{n} \frac{1}{\sqrt{2\pi\sigma^2}} exp\bigg(-\frac{(v_i-\mu_i)^2}{2\sigma^2_i}\bigg) \right]
\label{eq:ch:noncomp:nc:score:expanded:smooth}
\end{equation}

\begin{figure}[ht!]
\centering
 \begin{tikzpicture}[scale=0.9]
     \begin{axis}[
 	xlabel=$v_i$,
	ylabel=$v_j$,
	zlabel=$p(\mathbf{v})$,
 	yticklabels={,0,$\mu_j$,1},
	xticklabels={,0,,$\mu_i$,,1},
	 ]
     \addplot3[surf,fill=white,faceted color=blue!20!black,domain=-2:2] {exp(-x^2-y^2)};
	 \addplot3[surf,fill=green!95!blue,opacity=1,faceted color=green!50!blue,domain=0:2] {exp(-x^2-y^2)};
     \end{axis}
\end{tikzpicture}
\caption{A multivariate normal distribution with hypothetical variables $v_i$ and $v_j$. Assuming that $v_i$ and $v_j$ represent different characteristics of non-compositionality, most of non-compositional phrases are located in the highlighted region.}
\label{fig:ch:noncomp:multivar:normal:surf}
\end{figure}
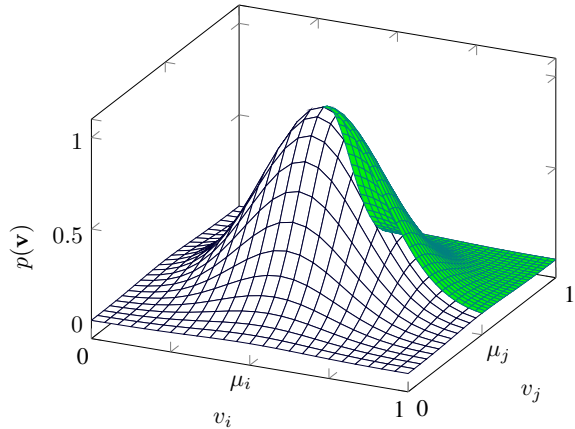

An illustration of the region where $\ncscore$ score is high in a bivariate distribution for hypothetical variables $v_i$ and $v_j$ is presented in Fig. \ref{fig:ch:noncomp:multivar:normal:surf}. The the region of interest that is most densely populated by non-compositional phrases is highlighted in green. As we move further from the $mean$ of this distribution towards the tail, the degree of non-compositionality of the phrases increases. $\ncscore$ and $\ncscoreSMOO$ guarantee to assign high scores to the phrases that are located on the tail of this distribution and increasingly lower the scores as we move toward the $mean$.

\section{Evaluation and Experiments}
\label{multivar:mul:model:eval}

In this section, we evaluate our core models i.e. $\ncscore$ and $\ncscoreSMOO$ on \dsfar{} and \dsred{} and compare their performance against $\additive$, $\compErrorInt$, and $\noncompmult$ baselines. 

\subsection{Evaluation on \dsfar{}}
\label{eval:far}

We calculate a vote-based score to represent the degree of non-compositionality, analogous to what was described in Sec. \ref{sec:identify:association}, but this time only based on the non-compositionality judgements. For evaluation on this dataset, we use $\pat k$ (see Sec. \ref{sec:eval:method} for more details). The performance of different models for different values of $k$ are shown in Fig. \ref{fig:dsfar}. We saw that $\additive$ and $\compErrorInt$ are two strong baselines from previous work that identify non-compositional phrases based only on their specific characteristic i.e. non-compositionality. $\noncompmult$ improves on the performance of these baselines implying the advantages of considering statistical association and non-substitutability as complementary pieces of information for the identification of non-compositional phrases. 

The multivariate models $\ncscore$ and $\ncscoreSMOO$ outperform $\noncompmult$ for most values of $k$ while they outperform $\additive$ and $\compErrorInt$ for almost all values of $k$ by a relatively large margin. $\ncscore$ performs better than $\ncscoreSMOO$ (the smooth version) at lower values of $k$, while $\ncscoreSMOO$ performs better at higher values of $k$. This implies that the smooth version has a higher recall. This is because $\ncscore$ assigns zero to any phrase for which at least one dimension had a value that is even slightly smaller than its corresponding $mean$ while $\ncscoreSMOO$ only slightly lowers the overall score of such phrase. On the other hand, evidently, in this dataset the value of at least one dimension for some of the non-compositional phrases fall below their corresponding $mean$. Such phrases are identified by $\ncscoreSMOO$ and hence it reaches a higher recall. 
\begin{figure}[ht!]
\centering
\begin{tikzpicture}

\pgfplotsset{grid style={help lines}}
\begin{axis}[
xmax=300,xmin=25,
ymax=0.7,ymin=0.30,
width=10cm,
height=6.5cm,
legend style={at={(1,1.1)},anchor=north east},
legend cell align=left,
xtick={50,100,150,200,250,300},
   xlabel={$k$},
   ylabel={$\pat k$},
    ylabel style = {font=\small,yshift=-2ex},
   yticklabel style = {font=\small,xshift=0.5ex},
   xticklabel style = {font=\small,yshift=0.5ex},
grid=both,
grid style={line width=.1pt, draw=gray!10},
major grid style={line width=.2pt,draw=gray!50},
minor tick num=1,
legend entries={$\additive$,$\compErrorInt$, $\noncompmult$, $\ncscore$,$\ncscoreSMOO$},
]

\addplot[color=purple,style=dashdotted,mark=star] table {data/ch_noncomp/baseline/additivek_pies.csv};

\addplot[color=red,mark=x] table {data/ch_noncomp/hybrid/hyb_normal_dist_k5_smooth/pk_error_avg.csv};

\addplot[color=orange,style=dashdotted,mark=star] table {data/ch_noncomp/hybrid/hyb_normal_dist_k5_smooth/pk_mult_avg.csv};

\addplot[color=green!80!black,mark=10-pointed star] table {data/ch_noncomp/hybrid/hyb_normal_dist_k5_smooth/pk_dist_avg.csv};

\addplot[color=blue,mark=x] table {data/ch_noncomp/hybrid/hyb_normal_dist_k5_smooth/pk_dist_smooth_avg.csv};

\end{axis}
\end{tikzpicture}
\caption{Performance of $\ncscore$ and $\ncscoreSMOO$ in comparison with the best previous models ($\additive$ and $\compErrorInt$) and $\noncompmult$, in terms of $\pat k$ on \dsfar{}.\label{fig:dsfar}}
\end{figure}

\subsection{Evaluation on \dsred{}}
\label{eval:red}
\begin{figure}[ht!]
\centering
\includegraphics[scale=0.39]{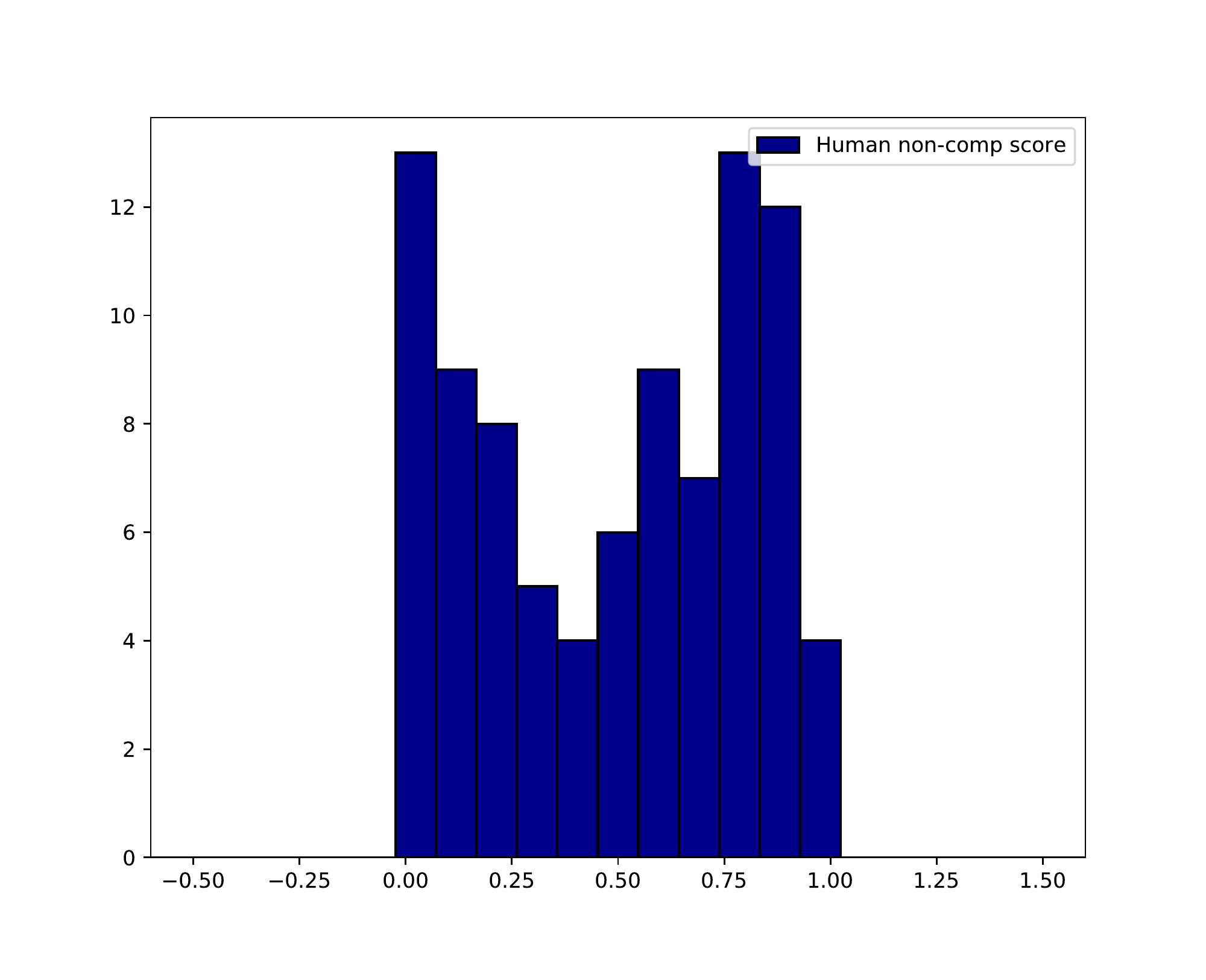}
\caption{Distribution of non-compositionality (inverse human compositionality) scores in \dsred{}. We expect to see a right-skewed distribution but ensuring that the dataset is balanced led to a clear bias that can be seen here.}
\label{reddy:noncomp:distribution}
\end{figure}

\begin{figure}[ht!]
\centering
\includegraphics[scale=0.36]{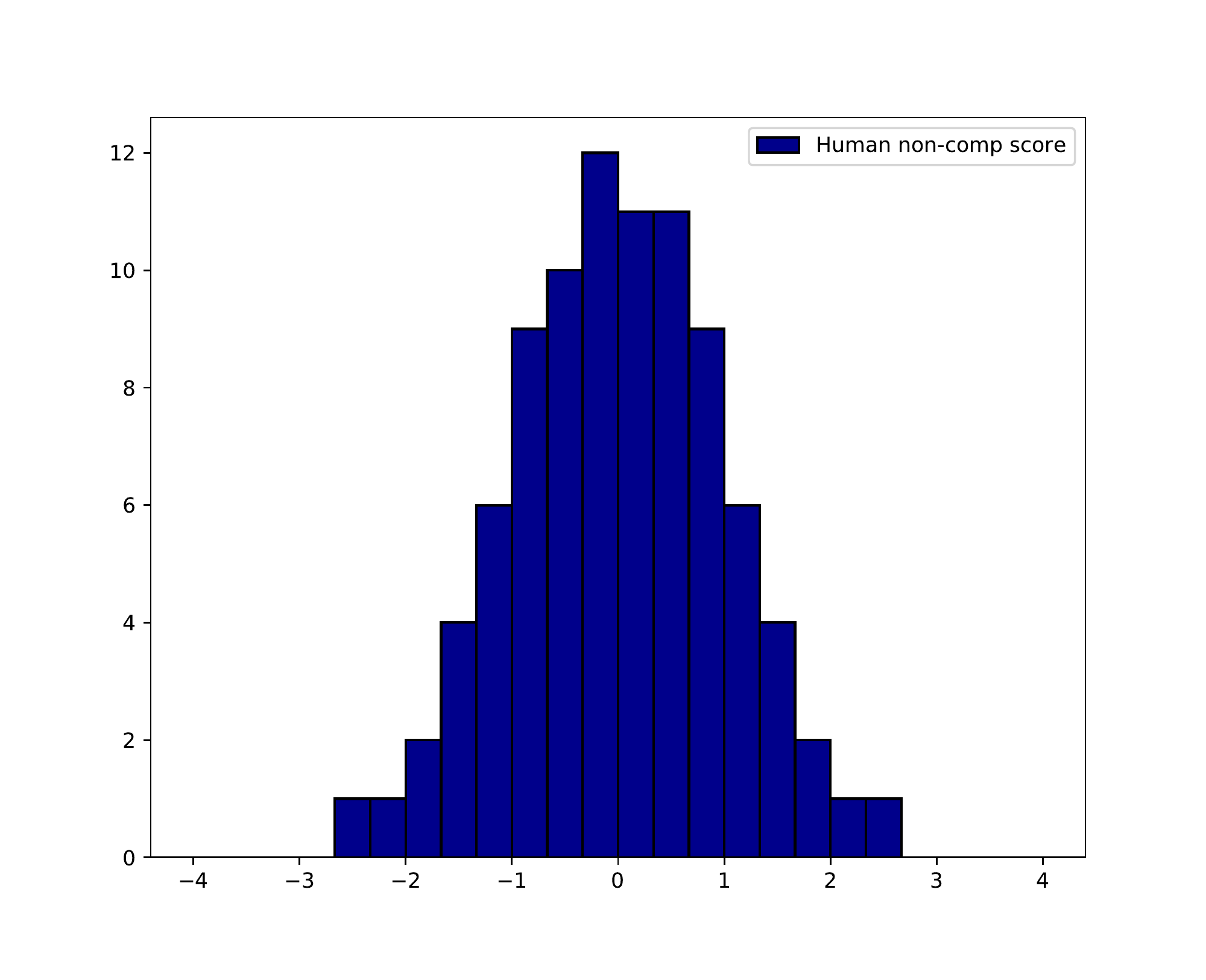}
\caption{Gaussianized distribution of non-compositionality (inverse human compositionality) scores in \dsred{}.}
\label{reddy:noncomp:distribution:normal}
\end{figure}

\begin{figure*}[!htbp]
\centering
  \subfloat[Human scores]{\includegraphics[scale=0.36]{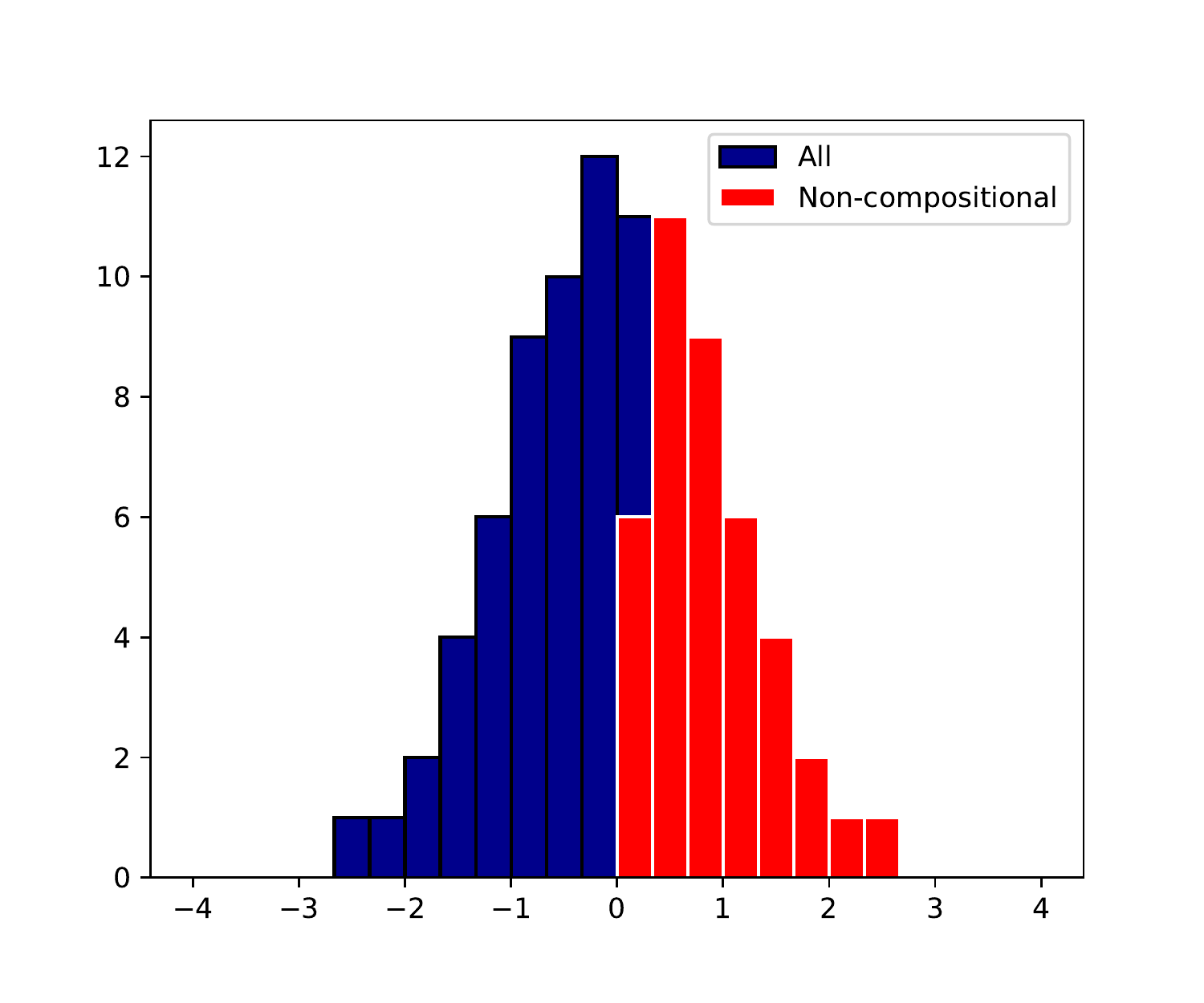}
  \label{fig:guassian:measures:reddy:1}
  }
  \subfloat[$\compErrorInt$]{\includegraphics[scale=0.36]{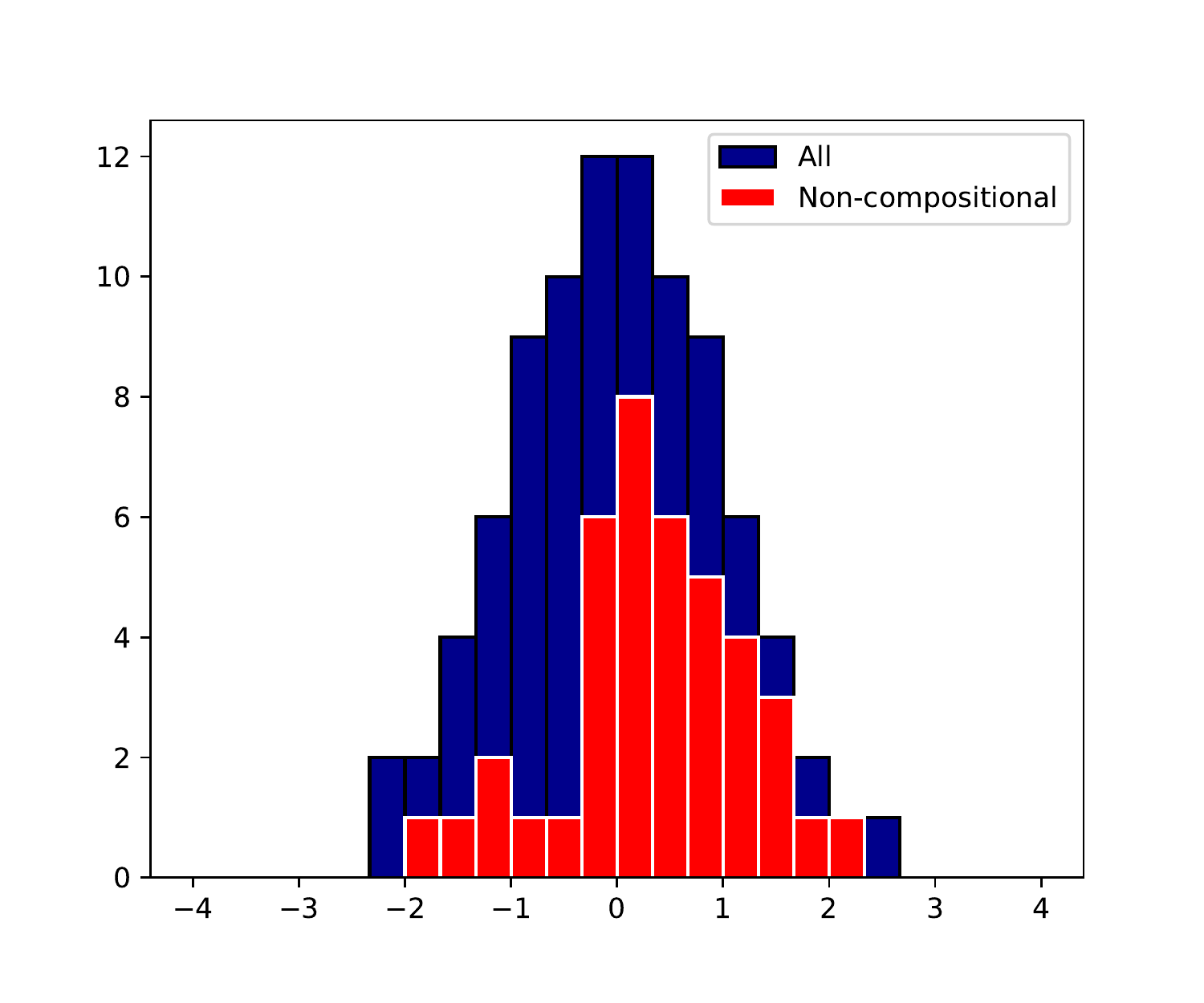}
  \label{fig:guassian:measures:reddy:2}
  }
  \subfloat[$\additive$]{\includegraphics[scale=0.36]{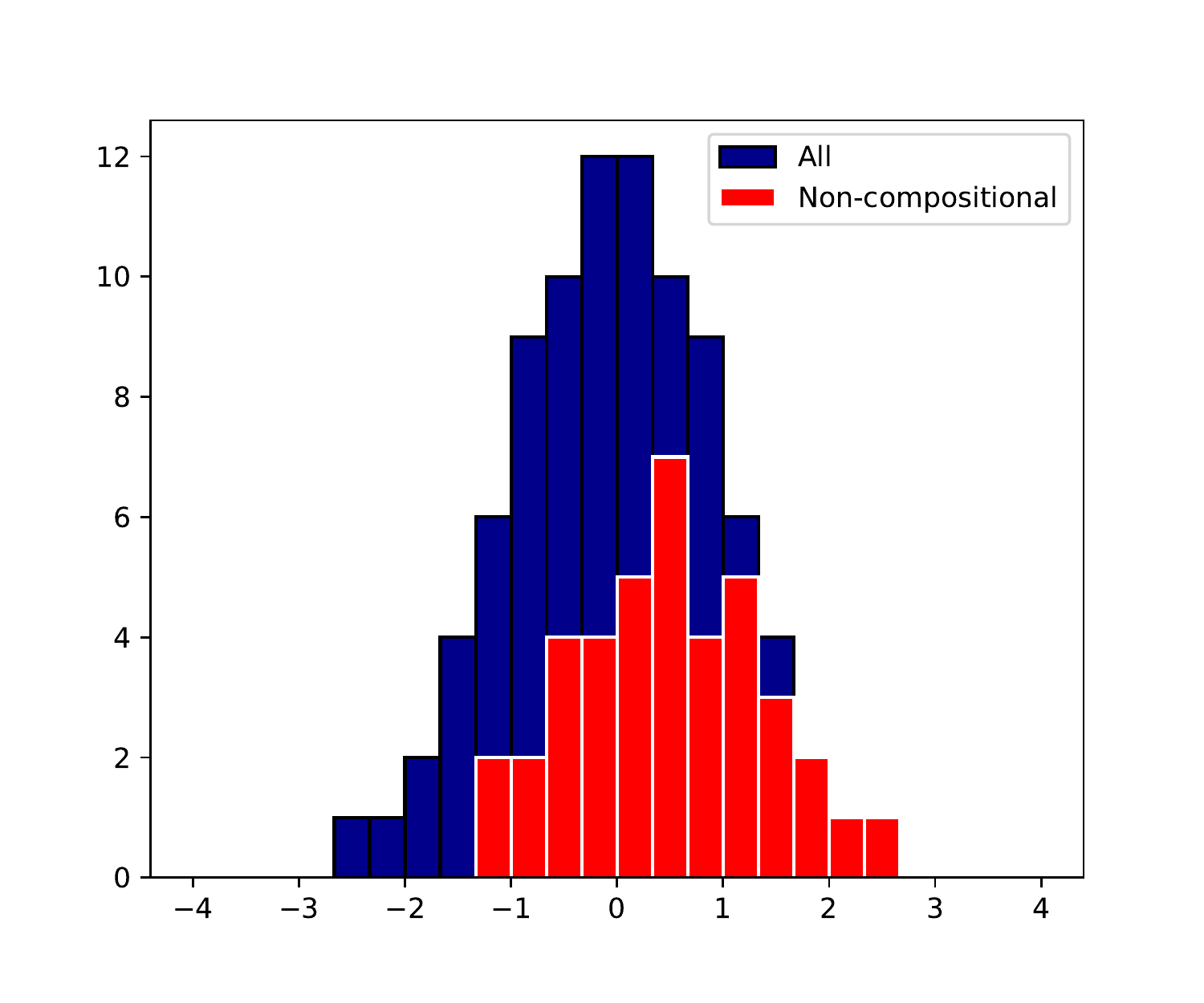}
  \label{fig:guassian:measures:reddy:3}
  }
  \vspace{-2mm}
  \\
  \subfloat[$\npmi$]{\includegraphics[scale=0.36]{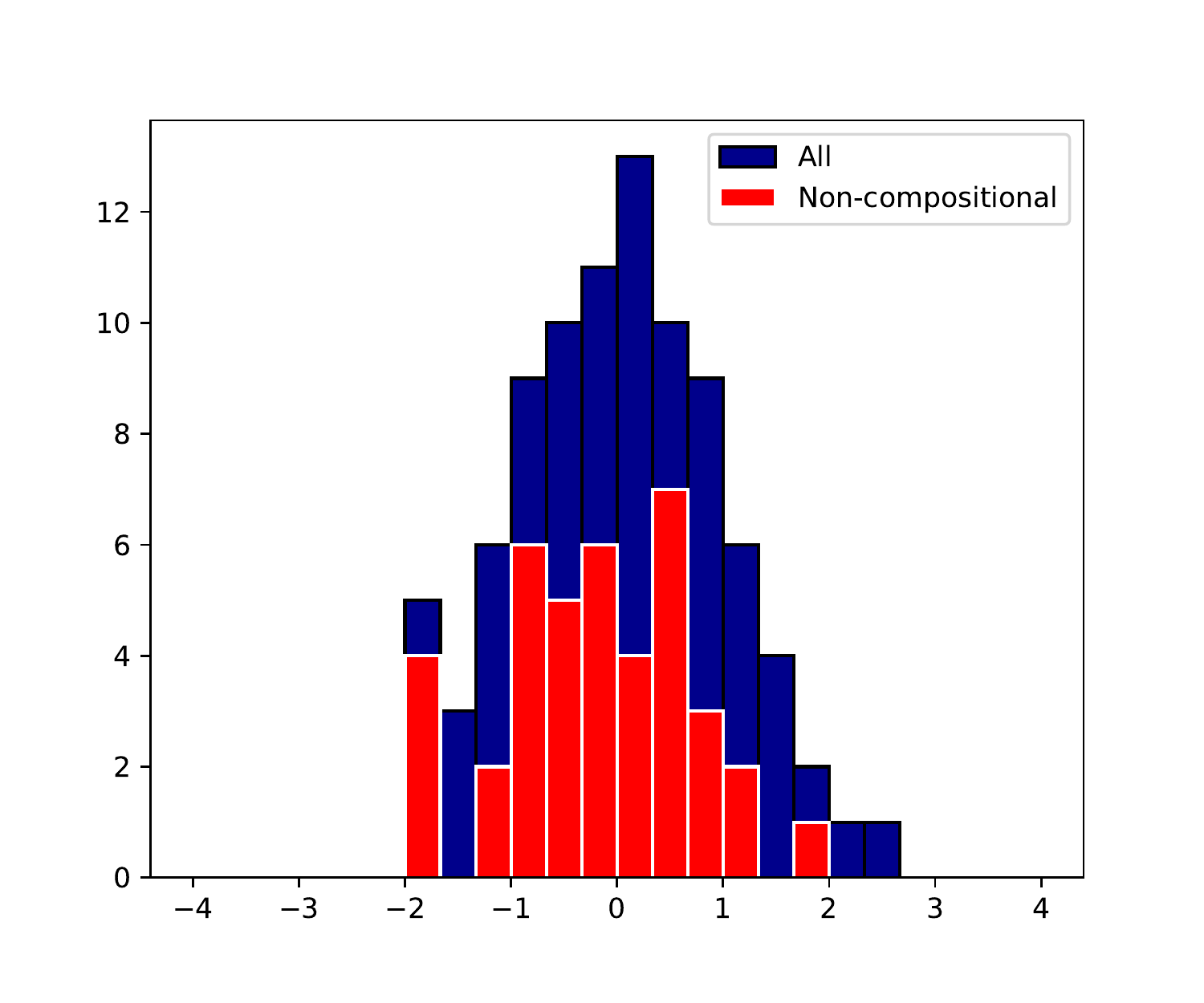}
  \label{fig:guassian:measures:reddy:4}
  }
  \subfloat[$\mOne$]{\includegraphics[scale=0.36]{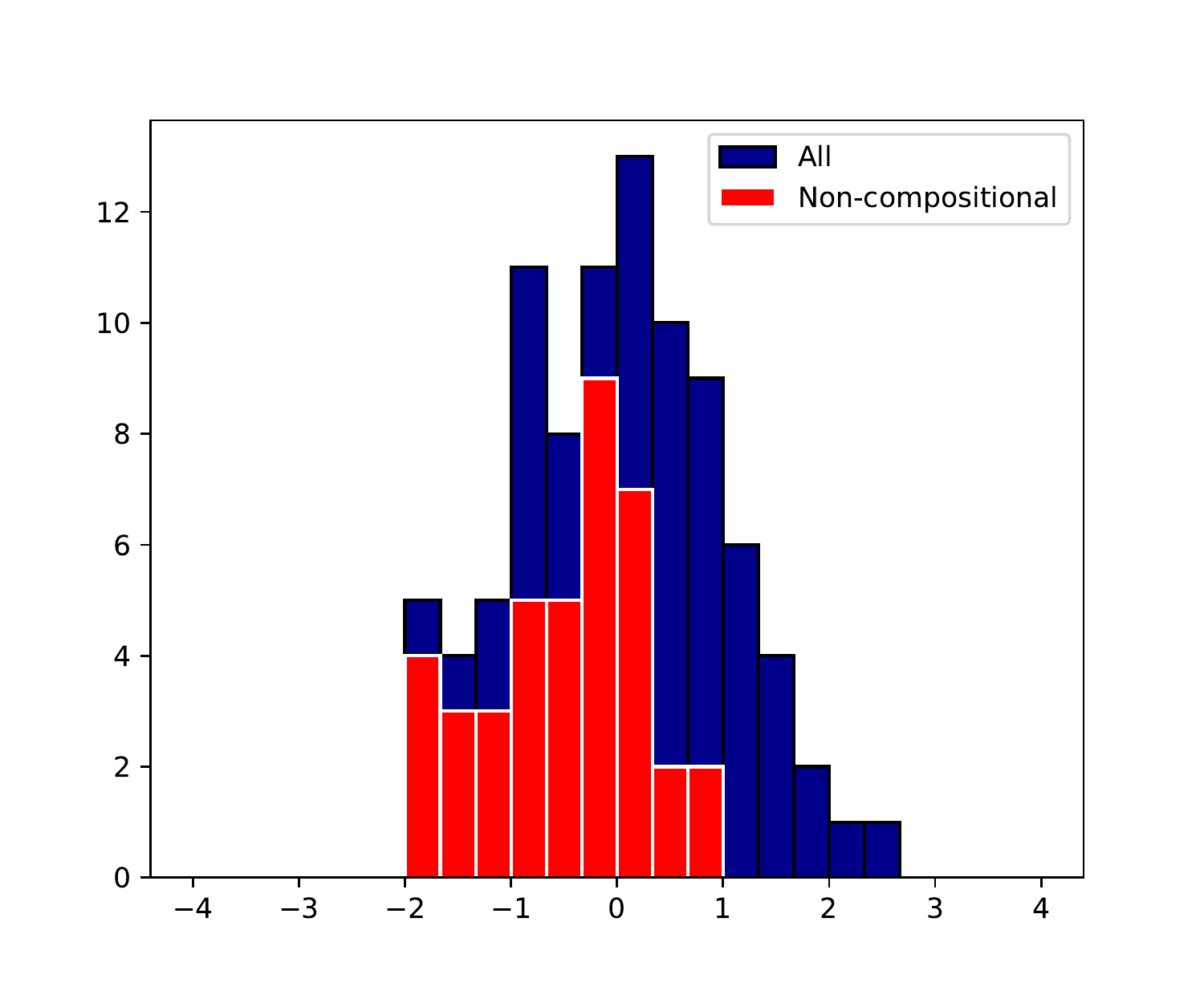}
  \label{fig:guassian:measures:reddy:5}
  }
\caption{Distribution of characteristic and human scores in \dsred{} after being transformed into normal.}
\label{fig:guassian:measures:reddy}
\vspace{-2mm}
\end{figure*}

\dsred{} comprises $90$ English compounds that are annotated with a compositionality score. Compositionality is defined as the property of a compound whose semantics is composed of the semantics of its components. They ask human judges to score (between $0$ and $5$) the compositionality of the compounds. We create the inverse of this score and regard it as non-compositionality score. For the dataset to incorporates sufficient number of non-compositional and compositional compounds, the compounds were selected manually and not through a random selection unlike the compounds of  \dsfar{}. But this led to an unnatural distribution where there are many non-compositional and compositional compounds and few compounds in between as  seen in Fig. \ref{reddy:noncomp:distribution}. This is often not expected when we look at the natural distribution of compositionality. We often expect to see a left-skewed distribution for a compositionality score and a right-skewed distribution for a non-compositionality score. Since the compounds of \dsred{} were not selected via random selection, they do not follow the expected distribution. Therefore, the multivariate model that is based on the normality assumption and natural distribution of the scores will have difficulty on this particular dataset. Nevertheless, we employ a complex transformation method i.e. the iterative method of \citet{5720319} to transform the distribution of the scores of this dataset to normal. Nevertheless, the above issue will not be eradicated. The result of the transformation can be seen in Fig. \ref{reddy:noncomp:distribution:normal}.
To illustrate how compositional vs non-compositional compounds are distributed, we assume any compound with a human non-compositionality score of above the $mean$ of human scores is non-compositional and it is compositional otherwise ({\it criterion A}). The normalized distribution of human non-compositionality scores, $\compErrorInt$, $\additive$, $\npmi$, and $\mOne$ are shown in Fig. \ref{fig:guassian:measures:reddy} in blue. For each distribution, the non-compositional compounds (derived with respect to {\it criterion A}) are highlighted in red.  

We evaluate the performance of different models on this dataset by means of Spearman's $\rho$. The correlations between human non-compositionality scores and the scores of different models is shown in Table \ref{table:correlations:on:reddy}. 
\begin{table}[]
\centering
\begin{tabular}{@{}lllll@{}}
\toprule
Model       	        & Spearman $\rho$ & p-value   &  &  \\ \midrule
$\additive$       & 0.387*** & 0.00016&  &  \\
$\compErrorInt$   & 0.433*** & 2.02e{-5}  &  &  \\
\bottomrule
$\npmi$ 		& -0.227	& 0.031 &	&\\
$\mOne$ 		& -0.555*** & 1.29e{-8} &	&\\
\bottomrule
\bottomrule
$\noncompmult$   & 0.473***         & 3.16e{-5}  &  &  \\ 
$\ncscoreSMOO$   & 0.530***         & 7.43e{-8}  &  &  \\ \bottomrule
\multicolumn{3}{l}{*** Significance at the 0.001 level}
\end{tabular}
\vspace{3mm}
\caption{Correlations bet. human non-compositionality scores and various characteristic scores in terms of Spearman $\rho$.}
\label{table:correlations:on:reddy}
\end{table}
As seen, $\npmi$ and $\mOne$ have a negative correlation with human non-compositionality scores. This is expected because as discussed earlier, the compounds of this dataset were selected ensuring that they are idiosyncratic at some level. Even the highly compositional compounds of this dataset (e.g. {\it bank account} and {\it end user}) have high statistical associations and hence, have a high $\npmi$ and SDMA score.
In other words, there are no (or very few) compounds that have a low statistical association or a low degree of non-substitutability in this dataset. Therefore, in this dataset, unlike in a randomly selected set of compounds, high association and non-substitutability are not discriminant features. This can also be confirmed by looking at Fig. \ref{fig:guassian:measures:reddy} where most non-compositional compounds are gathered in the right side of the distributions of $\additive$ and $\compErrorInt$, but in the left side of the distributions of $\npmi$ and $\mOne$. 

In theory, high $\npmi$ and $\mOne$ is a signal for non-compositionality, however, as discussed above, this is not the case in this particular dataset. Hence, we accordingly adjust $\noncompmult$ and $\ncscoreSMOO$ to incorporate only the scores with positive correlations with human scores i.e, $\additive$ and $\compErrorInt$. 
The performance of these two models in terms of Spearman $\rho$ correlation with human non-compositionality scores can be seen in the bottom section of Table \ref{table:correlations:on:reddy}. As seen, combining the {\em relevant} scores through multiplicative and multivariate model leads to an improved identification, such that the correlation with human judgements increases from $0.433$ (best previous baseline) to $0.473$ for $\noncompmult$ and $0.530$ for $\ncscoreSMOO$.

\section{Conclusions and Future Work}
\label{c_and_fw}
We defined the identification of non-compositional phrases as a multivariate problem, for which we presented a multivariate distribution-based model. We argued that taking into account general characteristics of non-compositional phrases, in addition to their non-compositionality, can considerably improve the identification of these phrases. We studied general and specific characteristics of non-compositional phrases and presented a multivariate model for their identification that takes into account all those characteristics. The proposed multivariate model can be expanded along different dimensions by taking into account other properties of non-compositional phrases, in future work. We focused only on noun compounds but in future work, the presented model can be applied to other syntactic categories of non-compositional MWEs. We made the assumption of independence between different characteristics of non-compositional phrases because they were measured by independent processes. In future work, the dependence and covariance between these characteristics can be studied and integrated into the multivariate model. Moreover, combining the mentioned characteristics through regression and similar techniques can be studied and compared with the proposed model. 

\bibliography{MWE_Bibdesk.bib}

\begin{thebibliography}{31}
\providecommand{\natexlab}[1]{#1}
\providecommand{\url}[1]{\texttt{#1}}
\expandafter\ifx\csname urlstyle\endcsname\relax
  \providecommand{\doi}[1]{doi: #1}\else
  \providecommand{\doi}{doi: \begingroup \urlstyle{rm}\Url}\fi

\bibitem[Berend(2011)]{berend2011opinion}
G{\'a}bor Berend.
\newblock Opinion expression mining by exploiting keyphrase extraction.
\newblock In \emph{IJCNLP}, pages 1162--1170. Citeseer, 2011.

\bibitem[Mitchell and Lapata(2008)]{mitchell2008vector}
Jeff Mitchell and Mirella Lapata.
\newblock Vector-based models of semantic composition.
\newblock In \emph{ACL}, 2008.

\bibitem[Reddy et~al.(2011)Reddy, McCarthy, and Manandhar]{reddy2011empirical}
Siva Reddy, Diana McCarthy, and Suresh Manandhar.
\newblock An empirical study on compositionality in compound nouns.
\newblock In \emph{IJCNLP}, pages 210--218, 2011.

\bibitem[Salehi et~al.(2015)Salehi, Cook, and Baldwin]{salehi2015embed}
Bahar Salehi, Paul Cook, and Timothy Baldwin.
\newblock A word embedding approach to predicting the compositionality of
  multiword expressions.
\newblock In \emph{Proceedings of NAACL HLT}. Association for Computational
  Linguistics, 2015.

\bibitem[Kiela and Clark(2013)]{kiela2013detecting}
Douwe Kiela and Stephen Clark.
\newblock Detecting compositionality of multi-word expressions using nearest
  neighbours in vector space models.
\newblock In \emph{EMNLP}, pages 1427--1432, 2013.

\bibitem[Lin(1999)]{Lin:1999}
Dekang Lin.
\newblock Automatic identification of non-compositional phrases.
\newblock In \emph{Proceedings of the 37th Annual Meeting of the Association
  for Computational Linguistics on Computational Linguistics}, ACL '99, pages
  317--324, Stroudsburg, PA, USA, 1999. Association for Computational
  Linguistics.
\newblock ISBN 1-55860-609-3.
\newblock \doi{10.3115/1034678.1034730}.
\newblock URL \url{http://dx.doi.org/10.3115/1034678.1034730}.

\bibitem[Tapanainen et~al.(1998)Tapanainen, Piitulainen, and
  J\"{a}rvinen]{Tapanainen:1998}
Pasi Tapanainen, Jussi Piitulainen, and Timo J\"{a}rvinen.
\newblock Idiomatic object usage and support verbs.
\newblock In \emph{Proceedings of the 36th Annual Meeting of the Association
  for Computational Linguistics and 17th International Conference on
  Computational Linguistics - Volume 2}, ACL '98, pages 1289--1293,
  Stroudsburg, PA, USA, 1998. Association for Computational Linguistics.
\newblock \doi{10.3115/980691.980779}.
\newblock URL \url{http://dx.doi.org/10.3115/980691.980779}.

\bibitem[Baldwin and Kim(2010)]{baldwin2010multiword}
Timothy Baldwin and Su~Nam Kim.
\newblock Multiword expressions.
\newblock In \emph{Handbook of Natural Language Processing, second edition.},
  pages 267--292. CRC Press, 2010.

\bibitem[Baldwin et~al.(2003)Baldwin, Bannard, Tanaka, and
  Widdows]{baldwin2003empirical}
Timothy Baldwin, Colin Bannard, Takaaki Tanaka, and Dominic Widdows.
\newblock An empirical model of multiword expression decomposability.
\newblock In \emph{Proceedings of the ACL 2003 workshop on Multiword
  expressions: analysis, acquisition and treatment-Volume 18}, pages 89--96.
  Association for Computational Linguistics, 2003.

\bibitem[McCarthy et~al.(2003)McCarthy, Keller, and
  Carroll]{McCarthy03detectinga}
Diana McCarthy, Bill Keller, and John Carroll.
\newblock Detecting a continuum of compositionality in phrasal verbs.
\newblock In \emph{Proceedings of the ACL-SIGLEX Workshop on Multiword
  Expressions: Analysis, Acqusation and Treatment}, pages 73--80, 2003.

\bibitem[Venkatapathy and Joshi(2005)]{venkatapathy:2005:measuring:VN}
Sriram Venkatapathy and Aravind~K. Joshi.
\newblock Measuring the relative compositionality of verb-noun (v-n)
  collocations by integrating features.
\newblock In \emph{Proceedings of the Conference on Human Language Technology
  and Empirical Methods in Natural Language Processing}, HLT '05, pages
  899--906, Stroudsburg, PA, USA, 2005. Association for Computational
  Linguistics.
\newblock \doi{10.3115/1220575.1220688}.
\newblock URL \url{http://dx.doi.org/10.3115/1220575.1220688}.

\bibitem[Katz and Giesbrecht(2006)]{Katz:2006}
Graham Katz and Eugenie Giesbrecht.
\newblock Automatic identification of non-compositional multi-word expressions
  using latent semantic analysis.
\newblock In \emph{Proceedings of the Workshop on Multiword Expressions:
  Identifying and Exploiting Underlying Properties}, MWE '06, pages 12--19,
  Stroudsburg, PA, USA, 2006. Association for Computational Linguistics.
\newblock ISBN 1-932432-84-1.
\newblock URL \url{http://dl.acm.org/citation.cfm?id=1613692.1613696}.

\bibitem[Hermann et~al.(2012)Hermann, Blunsom, and
  Pulman]{hermann2012unsupervised}
Karl~Moritz Hermann, Phil Blunsom, and Stephen Pulman.
\newblock An unsupervised ranking model for noun-noun compositionality.
\newblock In \emph{Proceedings of the First Joint Conference on Lexical and
  Computational Semantics}, pages 132--141. Association for Computational
  Linguistics, 2012.

\bibitem[Schulte Im~Walde et~al.(2013)Schulte Im~Walde, M{\"u}ller, and
  Roller]{im2013exploring}
Sabine Schulte Im~Walde, Stefan M{\"u}ller, and Stephen Roller.
\newblock Exploring vector space models to predict the compositionality of
  german noun-noun compounds.
\newblock In \emph{Proceedings of the 2nd Joint Conference on Lexical and
  Computational Semantics}, pages 255--265, 2013.

\bibitem[Yazdani et~al.(2015)Yazdani, Farahmand, and
  Henderson]{yazdaniFarahmand15}
Majid Yazdani, Meghdad Farahmand, and James Henderson.
\newblock Learning semantic composition to detect non-compositionality of
  multiword expressions.
\newblock In \emph{Proceedings of the 2015 Conference on Empirical Methods in
  Natural Language Processing}, pages 1733--1742, Lisbon, Portugal, September
  2015. Association for Computational Linguistics.
\newblock URL \url{http://aclweb.org/anthology/D15-1201}.

\bibitem[Cordeiro et~al.(2016)Cordeiro, Ramisch, Idiart, and
  Villavicencio]{P16-1187}
Silvio Cordeiro, Carlos Ramisch, Marco Idiart, and Aline Villavicencio.
\newblock Predicting the compositionality of nominal compounds: Giving word
  embeddings a hard time.
\newblock In \emph{Proceedings of the 54th Annual Meeting of the Association
  for Computational Linguistics (Volume 1: Long Papers)}, pages 1986--1997.
  Association for Computational Linguistics, 2016.
\newblock \doi{10.18653/v1/P16-1187}.
\newblock URL \url{http://aclweb.org/anthology/P16-1187}.

\bibitem[Farahmand et~al.(2015)Farahmand, Smith, and Nivre]{farahmand2015data}
Meghdad Farahmand, Aaron Smith, and Joakim Nivre.
\newblock A multiword expression data set: Annotating non-compositionality and
  conventionalization for english noun compounds.
\newblock In \emph{Proceedings of the 11th Workshop on Multiword Expressions
  (MWE-NAACL 2015)}. Association for Computational Linguistics, 2015.

\bibitem[Evert(2005{\natexlab{a}})]{evert2005statistics}
Stefan Evert.
\newblock \emph{The statistics of word cooccurrences}.
\newblock PhD thesis, Stuttgart University, 2005{\natexlab{a}}.

\bibitem[Evert(2005{\natexlab{b}})]{evert2005statistics-patk}
Stefan Evert.
\newblock \emph{The statistics of word cooccurrences}, chapter~5, pages
  138--140.
\newblock 2005{\natexlab{b}}.

\bibitem[Church and Hanks(1990)]{church1990word}
Kenneth~Ward Church and Patrick Hanks.
\newblock Word association norms, mutual information, and lexicography.
\newblock \emph{Computational linguistics}, 16\penalty0 (1):\penalty0 22--29,
  1990.

\bibitem[Bouma(2009)]{bouma2009normalized}
G.~Bouma.
\newblock {Normalized (pointwise) mutual information in collocation
  extraction}.
\newblock In \emph{From Form to Meaning: Processing Texts Automatically,
  Proceedings of the Biennial GSCL Conference 2009}, volume Normalized, pages
  31--40, T\"{u}bingen, 2009.

\bibitem[Dunning(1993)]{dunning1993accurate}
Ted Dunning.
\newblock Accurate methods for the statistics of surprise and coincidence.
\newblock \emph{Computational linguistics}, 19\penalty0 (1):\penalty0 61--74,
  1993.

\bibitem[Smadja et~al.(1996)Smadja, McKeown, and
  Hatzivassiloglou]{smadja1996translating}
Frank Smadja, Kathleen~R McKeown, and Vasileios Hatzivassiloglou.
\newblock Translating collocations for bilingual lexicons: A statistical
  approach.
\newblock \emph{Computational linguistics}, 22\penalty0 (1):\penalty0 1--38,
  1996.

\bibitem[Acosta et~al.(2011)Acosta, Villavicencio, and
  Moreira]{acosta2011identification}
Otavio Acosta, Aline Villavicencio, and Viviane Moreira.
\newblock Identification and treatment of multiword expressions applied to
  information retrieval.
\newblock \emph{Kordoni et al}, pages 101--109, 2011.

\bibitem[Schone and Jurafsky(2001)]{schone2001knowledge}
Patrick Schone and Daniel Jurafsky.
\newblock Is knowledge-free induction of multiword unit dictionary headwords a
  solved problem.
\newblock In \emph{Proceedings of the 2001 Conference on Empirical Methods in
  Natural Language Processing}, pages 100--108, 2001.

\bibitem[Pecina(2010)]{pecina}
Pavel Pecina.
\newblock Lexical association measures and collocation extraction.
\newblock \emph{Language resources and evaluation}, 44\penalty0 (1-2):\penalty0
  137--158, 2010.

\bibitem[Manning and Sch{\"u}tze(1999)]{manning1999foundations}
Christopher~D Manning and Hinrich Sch{\"u}tze.
\newblock \emph{Foundations of statistical natural language processing}.
\newblock MIT press, 1999.

\bibitem[Pearce(2001)]{pearce2001synonymy}
Darren Pearce.
\newblock Synonymy in collocation extraction.
\newblock In \emph{Proceedings of the Workshop on WordNet and Other Lexical
  Resources, Second meeting of the North American Chapter of the Association
  for Computational Linguistics}, pages 41--46. Citeseer, 2001.

\bibitem[Ramisch(2012)]{ramisch2012generic}
Carlos Ramisch.
\newblock A generic framework for multiword expressions treatment: from
  acquisition to applications.
\newblock In \emph{Proceedings of ACL 2012 Student Research Workshop}, pages
  61--66. Association for Computational Linguistics, 2012.

\bibitem[Farahmand and Henderson(2016)]{frahmand2016loglinear}
Meghdad Farahmand and James Henderson.
\newblock Proceedings of the 12th workshop on multiword expressions.
\newblock Association for Computational Linguistics, 2016.
\newblock \doi{10.18653/v1/W16-1809}.

\bibitem[Laparra et~al.(2011)Laparra, Camps-Valls, and Malo]{5720319}
V.~Laparra, G.~Camps-Valls, and J.~Malo.
\newblock Iterative gaussianization: From ica to random rotations.
\newblock \emph{IEEE Transactions on Neural Networks}, 22\penalty0
  (4):\penalty0 537--549, April 2011.
\newblock ISSN 1045-9227.
\newblock \doi{10.1109/TNN.2011.2106511}.

\end{thebibliography}
\bibliographystyle{unsrtnat}

\end{document}